\documentclass[10pt,twocolumn,letterpaper]{article}

\usepackage[pagenumbers]{cvpr} 

\usepackage{graphicx}
\usepackage{amsmath}
\usepackage{amssymb}
\usepackage{booktabs}
\usepackage{multirow}
\usepackage{placeins}
\usepackage{acronym}
\usepackage{bbold}
\usepackage{color}
\usepackage{verbatim}
\usepackage{xcolor}

\newcommand{\realn}{{\mathbb{R}}}
\newcommand{\norm}[1]{\lVert #1 \rVert}

\definecolor{cvprblue}{rgb}{0.21,0.49,0.74}
\usepackage[pagebackref,breaklinks,colorlinks,allcolors=cvprblue]{hyperref}

\usepackage[capitalize]{cleveref}
\crefname{section}{Sec.}{Secs.}
\Crefname{section}{Section}{Sections}
\Crefname{table}{Table}{Tables}
\crefname{table}{Tab.}{Tabs.}

\def\paperID{****} 
\def\confName{****}
\def\confYear{****}

\title{Prompt-Guided Attention Head Selection for Focus-Oriented Image Retrieval}

\author{Yuji Nozawa \qquad Yu-Chieh Lin \qquad Kazumoto Nakamura \qquad Youyang Ng \\
Kioxia Corporation \\
\small{\texttt{\{yuji1.nozawa,yuchieh.lin,kazumoto1.nakamura,youyang.ng\}@kioxia.com}}
}

\begin{document}
\maketitle

\begin{abstract}
  The goal of this paper is to enhance pretrained Vision Transformer (ViT) models for focus-oriented image retrieval with visual prompting. In real-world image retrieval scenarios, both query and database images often exhibit complexity, with multiple objects and intricate backgrounds. Users often want to retrieve images with specific object, which we define as the Focus-Oriented Image Retrieval (FOIR) task. While a standard image encoder can be employed to extract image features for similarity matching, it may not perform optimally in the multi-object-based FOIR task. This is because each image is represented by a single global feature vector. To overcome this, a prompt-based image retrieval solution is required. We propose an approach called Prompt-guided attention Head Selection (PHS) to leverage the head-wise potential of the multi-head attention mechanism in ViT in a promptable manner. PHS selects specific attention heads by matching their attention maps with user's visual prompts, such as a point, box, or segmentation. This empowers the model to focus on specific object of interest while preserving the surrounding visual context. Notably, PHS does not necessitate model re-training and avoids any image alteration. Experimental results show that PHS substantially improves performance on multiple datasets, offering a practical and training-free solution to enhance model performance in the FOIR task.
\end{abstract}

\section{Introduction}
\label{sec:intro}

\begin{figure}[t]
  \centering
  \includegraphics[width=0.95\linewidth]{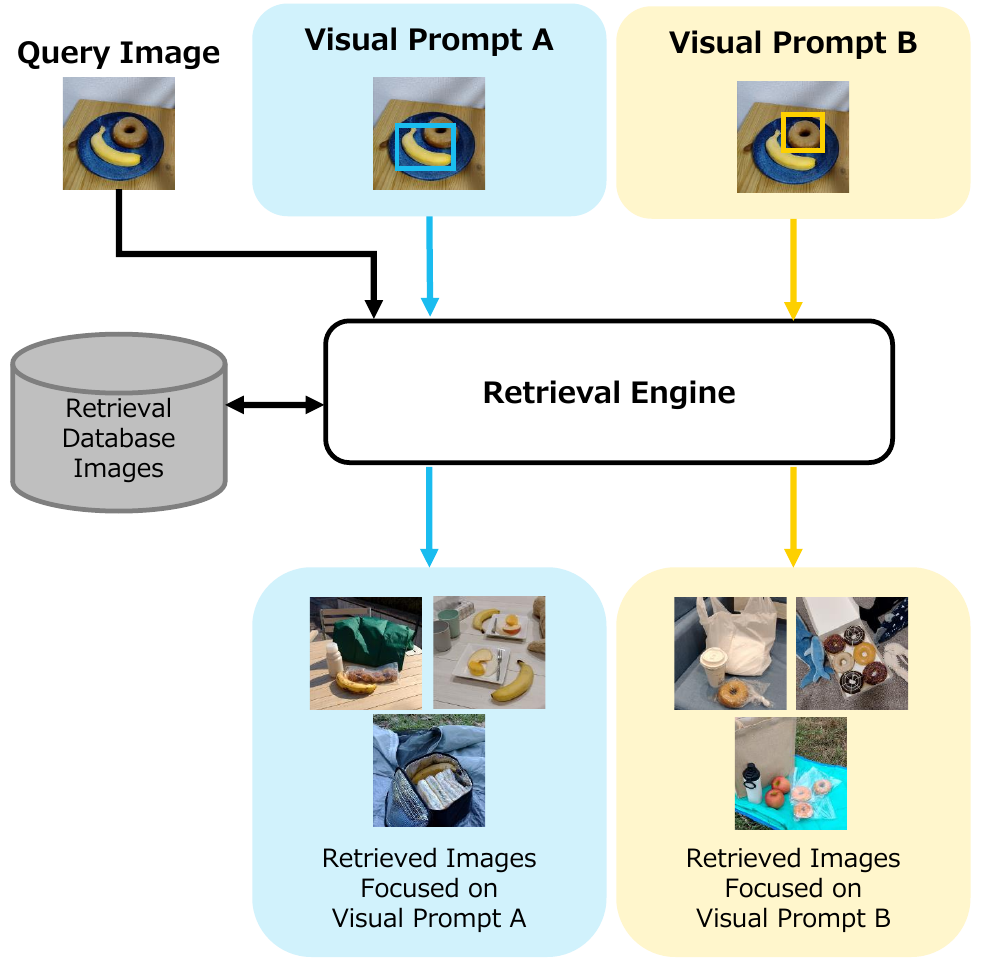}
  \caption{Overview of Focus-Oriented Image Retrieval (FOIR) task with illustration of visual prompting approach. FOIR task simulates real-world scenarios wherein (1) both query and retrieval database images often exhibit complexity, with multiple objects and intricate backgrounds; (2) users are visually interested in retrieving images containing specific object.}
  \label{fig:overviewtask}
\end{figure}

Image retrieval (IR) encompasses a wide range of applications, including face recognition~\cite{Schroff_2015_CVPR}, landmark retrieval~\cite{iccv2017largescale}, and online shopping~\cite{iccv2015wheretobyit}. A common approach for retrieving images involves extracting image features and determining their similarity. This process is referred to as Content-Based Image Retrieval (CBIR)~\cite{smeulders2000content,eccv2016deep,iccv2017largescale}. Standard CBIR approach demonstrates good performance when dealing with less complicated images, especially those that contain a single object. To improve the accuracy, researchers have employed deep neural network models such as Convolutional Neural Networks~\cite{NIPS2012_c399862d} and Vision Transformers (ViTs)~\cite{dosovitskiy2021an} to extract image features, and have localized the retrieval problem to specific objects, a task setting known as Localized CBIR~\cite{localizedCBIR} or Object-Based Image Retrieval (OBIR)~\cite{objectIR}. However, existing studies in Localized CBIR often employ image alteration preprocessing techniques such as masking and cropping, optimize the retrieval of images with less complexity, and tend to overlook the consideration of user intention~\cite{localizedCBIR,objectIR,regionhybrid,MSIR}. In reality, users may have the desire to retrieve a specific object while considering certain visual contextual constraints in complex images. The use of image alteration techniques can introduce errors during preprocessing or result in a loss of visual context~\cite{falip}, leading to retrieval failures. These inherent limitations potentially hinder its practical applicability.

In real-world image retrieval, both the query image and the images in the retrieval database often exhibit complexity, characterized by the presence of multiple objects and intricate backgrounds, posing a challenge for users aiming to retrieve images with specific objects. Despite its practicality, this topic has received limited research attention thus far. We argue that this specific scenario warrants dedicated attention in the implementation of image retrieval systems. In light of this, we set up Focus-Oriented Image Retrieval (FOIR) task, specifically for retrieving objects from complex images, as demonstrated in \cref{fig:overviewtask}. FOIR can be considered a distinct category within the Localized CBIR task, wherein both the query and retrieval database consist of complex images, and users are interested in retrieving images with specific object. Query formats can be classified into two main categories: whole-image-as-query (WIQ) and image-region-as-query (IRQ)~\cite{irq}. FOIR task specifically addresses WIQ, wherein the entire image is utilized to conduct the query. One solution is to employ a standard image encoder to extract image features for similarity matching. However, this may not work well for multi-object-based FOIR tasks as it represents each image with a single global feature vector. Image encoders tend to focus on the most salient region, potentially ignoring other objects or regions of interest. Prompt-based preprocessing techniques such as image cropping can complement the image encoder but may fail when wider visual context is required, as the perceptions of the user and the model may not align. Therefore, a solution that \textit{considers user visual preference, preserves visual context, and aligns user-model perceptions} is essential. 

To overcome this, we propose leveraging the head-wise potential of the Multi-Head Attention (MHA) mechanism of ViT in a user promptable manner. The attention heads in ViT, particularly in the last layers, contain valuable high-level salient and segmentation information~\cite{DINObib} for images. This information can be effectively utilized to enable object focusing in FOIR task. Attention maps from each head have been observed to focus on different objects or parts in an input image~\cite{DINObib}. Building on this observation, we introduce Prompt-guided attention Head Selection (PHS) technique. PHS selects specific attention heads in the last layer of a pretrained ViT image encoder model by matching their attention maps with user's visual prompts, which can take the form of a point, box, or segmentation. This empowers the model to focus on specific object of interest while preserving the surrounding visual context. Notably, PHS does not require model re-training, making it a plug-and-play enhancement for off-the-shelf pretrained ViT models. Moreover, our method avoids any modifications to the input images, thereby avoiding any undesirable consequences of image editing. Through extensive experimental evaluations on multiple datasets, we demonstrate that PHS substantially improves performance and robustness by selecting attention heads that are more focused on the desired objects.

From a conceptual standpoint, our proposed method can be regarded as a high-level perception matching mechanism between human users and vision models. A recent work of Foveal Attention (FA)~\cite{falip}, which draws inspiration from human visual perception~\cite{BURT2021145,peripheryfovea}, has endeavored to manipulate attention through visual prompting by incorporating attention mask into region of attention via addictive or blending operation. However, this approach inevitably alters the perception of attention heads themselves, favoring the user's perception over that of the vision model. In contrast, our method selects attention heads to align and match the user's perceptions with specific attention heads, thereby bridging the gap between human and model visual understanding without explicitly modifying the attentions of individual heads. Our method can also be viewed as grounding high-level human perception, in the form of visual prompting, to high-level perception that can be comprehended by the model, in the form of attention map in each head. Both FA and our method differ conceptually and possess their own strengths. FA aims to emulate human attention characteristics, while PHS strives to match human-model attentions. We believe that they are effective in their own right and also have a complementary relationship in bridging human-model perceptions.

Our contributions are summarized as follows:
\begin{itemize}
  \item	We design Prompt-guided attention Head Selection (PHS), a method that leverages the head-wise potential of the multi-head attention mechanism in ViT in a promptable manner to accommodate user visual preference, preserve visual context and align user-model perceptions.
  \item We empirically show that our proposed method enhances performance and robustness compared to existing methods across multiple datasets through extensive experiments and ablation studies in the Focus-Oriented Image Retrieval (FOIR) task, where both query and retrieval database images often exhibit complexity, with multiple objects and intricate backgrounds, and users are visually interested in retrieving images with specific object.
\end{itemize}

\section{Related Work}
\label{sec:related}
\subsection{Object-Based IR with User Interest}
\label{subsec:objectbased}
OBIR~\cite{objectIR,localizedCBIR,OESIR} localizes objects in retrieval task. As a solution, Ref.~\cite{regionhybrid} proposes image alterations to improve retrieval performance, while Ref.~\cite{aggregated} applies aggregated features for image retrieval. Ref.~\cite{MOIR} approaches multi-object IR from a CBIR perspective. MSIR~\cite{MSIR} sets up multi-subject IR based on complex images, similar to our task. However, it does not take into account user interest. In comparison, FOIR task can be seen as a specific category of OBIR, emphasizing on complex images and considering user visual intention. Several works have incorporated user intention into image retrieval using techniques such as sketch~\cite{cvpr2016sketch,Yelamarthi_2018_ECCV,Ghosh_2019_ICCV}, Composed IR with text prompts~\cite{NEURIPS2018_a01a0380,Guo2019TheFI,Vo_2019_CVPR,Chen_2020_CVPR,eccv2020learningjoint,Lee_2021_CVPR,wacv2021compositional,wacv2023fashion}, and relevancy~\cite{nara2024revisitrf}, but these task setups and methods require additional feedback data. In contrast, our task setup considers visual intention in query image, and our method employs visual prompting, which is a less demanding approach.

\begin{figure*}[t]
  \centering
  \includegraphics[width=\textwidth]{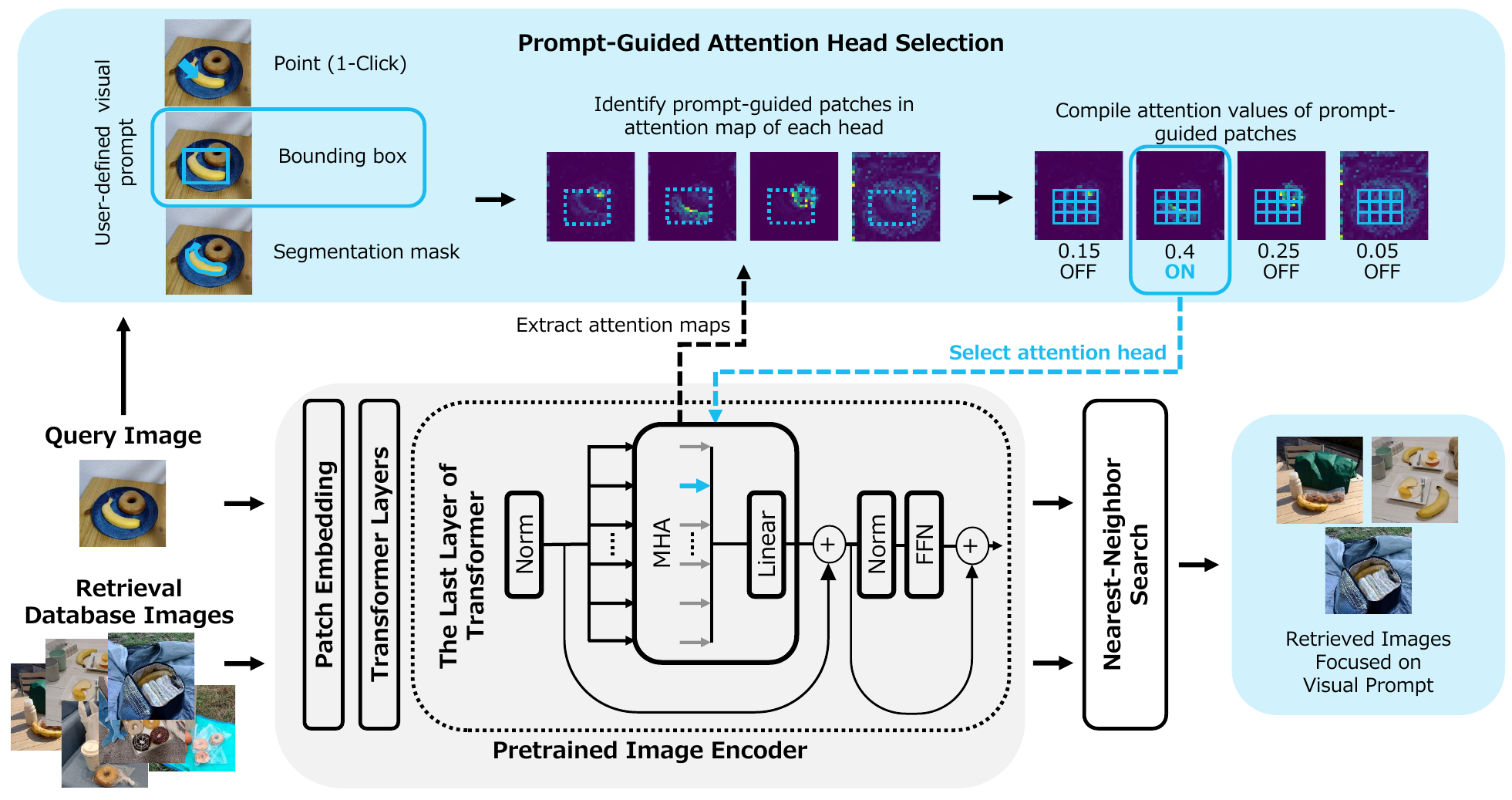}
  \caption{Overview of Prompt-guided attention Head Selection (PHS). PHS performs the selection of transformer attention heads in the pretrained ViT model by matching their attention maps with user's visual prompt, which can take the form of a point, box, or segmentation. This empowers the model to focus on specific object of interest while preserving the surrounding visual context. Best viewed in color.}
  \label{fig:overviewmethod}
\end{figure*}

\subsection{Visual Region Awareness \& Prompting in ViT}
\label{subsec:visualprompt}
ViT~\cite{dosovitskiy2021an} is a transformer-based~\cite{NIPS2017_3f5ee243} model for Computer Vision (CV) tasks. Transformer architectures use self-attention mechanism in MHA module to capture complex relationships between tokens~\cite{NIPS2017_3f5ee243}. DINO~\cite{DINObib} is one of the self-supervised learning (SSL) methods~\cite{DINObib,xie2021selfsupervised,Chen_2021_ICCV,li2022efficient} applied to ViT, shown to have work well in various downstream tasks. DINOv2~\cite{oquab2023dinov2} further improves its performance by pretraining on a large-scale visual dataset. DINO and DINOv2 have shown to exhibit beneficial region and object awareness in the MHA module of the last layer~\cite{simeoni2021localizing,wang2022self,wang2023cut}. Based on this characteristic, we primarily employ DINOv2 in our study but note that our method can generalize to models with different pretraining approach.

Region awareness is crucial in CV tasks like object detection and segmentation. Studies have enhanced CLIP~\cite{regionclip,sideadapter} and ViT models~\cite{chen2022regionvit,deformable,swintransformer} to incorporate region awareness mechanisms. However, these methods usually require additional training, while our method focuses on inference-time attention manipulation through user-defined visual prompting. Similar to prompt engineering~\cite{gpt} in Natural Language Processing (NLP), various visual prompts have been explored in CV, such as points, boxes \& masks~\cite{segmentanything}, blurred surroundings~\cite{finegrain}, red circles~\cite{redcircle}, foveal attention~\cite{falip} and learnable prompts~\cite{prompttuning,fairvpt,promptself,bahng2022visual}. However, unlike our method, most of them require input image alterations. FALIP~\cite{falip}'s approach is most relevant to our study, but it directly modifies individual attention heads while our method prioritizes user-model perception matching. Our method also enables multiple prompts types, similar to SAM~\cite{segmentanything}. In addition, these existing works did not specifically examine the area of visual-prompt-guided image retrieval, which remains underexplored. Our work aims to contribute insights into the intersection of visual prompts, user-model perceptions, and image retrieval.

\subsection{Attention Manipulation \& Head Selection}
\label{subsec:hs}
Numerous studies have investigated changes to the attention mechanism in ViT. Some have modified attention during training~\cite{chen2023accumulated,lee2021vision}, while others have focused on manipulating attention during inference~\cite{falip,itae,zhang2024personalize}. However, few studies have specifically examined attention manipulation techniques without altering individual attention values like head selection. Head selection is a technique used in the MHA module of Transformers in NLP and CV domains. It prioritizes relevant heads and replaces less relevant heads with fixed values. In NLP, it improves multilingual machine translation by selecting heads based on languages being translated~\cite{gong2021pay}. In CV, it enhances inference efficiency~\cite{Meng_2022_CVPR,MHSTbib}, knowledge distillation~\cite{pmlr-v162-wu22c}, and generalization~\cite{nicolicioiu2023learning}. Head selection in these previous works are performed and optimized for each task through training. Conversely, our work introduces inference-time head selection with a pretrained ViT model for image retrieval using visual prompting. The selection of heads is not fixed in advance through training. To the best of our knowledge, this is the first attempt to apply head selection to image retrieval.

\section{Approach}
\label{sec:approach}

In this section, we introduce our task setup (\cref{fig:overviewtask}) and proposed method (\cref{fig:overviewmethod}).

\subsection{Focus-Oriented Image Retrieval Task}
\label{subsec:foir}

We set up Focus-Oriented Image Retrieval task with the objective of simulating real-world image retrieval scenarios. FOIR can be considered a distinct category within the Localized CBIR tasks. The FOIR task encompasses the following criteria: (1) Image-to-image retrieval; (2) The query and retrieval database images exhibit complexity, characterized by the presence of multiple objects and intricate backgrounds; (3) Users are visually interested in retrieving images with specific object or pattern. For query format, FOIR task specifically addresses whole-image-as-query (WIQ), wherein the entire image is utilized to conduct the query. An exemplary demonstration of the FOIR task is the vision of a general-purpose humanoid robot. In this scenario, a human user may visually request the robot's assistance in searching for objects similar to what it is currently perceiving. The search can occur within the robot's past or future visual memory of its daily routine. Both the visual memory and the current vision of the real-world environment exhibit complexity with multiple objects and intricate backgrounds in each visual input.

In FOIR task, to accommodate user preference, a prompt-based image retrieval solution is essential. \cref{fig:overviewtask} illustrates an example of the FOIR task with a visual prompting approach, where the user creates visual prompts (A or B) to enhance the retrieval of images of interest from a complex database. In order to prevent retrieval failures caused by query preprocessing errors or a reduction in visual context, it is essential to use a non-image-alteration technique that preserves the available visual context. However, the presence of rich visual context can negatively impact retrieval results unless the user's and model's perceptions are aligned. Therefore, it is crucial to have a technique that aligns user and model perceptions. Overall, to provide a robust solution to the FOIR task, we have devised a method that (1) considers user's visual preference, (2) preserves visual context, and (3) aligns user and model perceptions. 

\subsection{Retrieval Process}
\label{subsec:retrieval}

We first introduce the overall image retrieval pipeline without head selection in this section. We employ pretrained ViT model to extract features of images. Let $\mathbf{x}_{\mathrm{input}}\in\realn^{H\times W\times C}$ be an input image, where $H$, $W$, and $C$ represent the height, width, and channel of the image respectively. The image is divided into 2D patches $\mathbf{x}_{\mathrm{patch}}\in \realn^{N\times P\times P\times C}$, where the size of each patch is $P\times P\times C$, and $N=HW/P^2$ is the number of patches. After the patches are embedded by a trained linear projection layer, the \verb|[CLS]| token is added to the patch tokens, and then the positional encoding is added to each embedding token. Consequently, $\mathbf{x}_{\mathrm{patch}}$ is transformed into $\mathbf{x}_{\mathrm{token}}\in \realn^{(N+1)\times d}$, where $d$ denotes the embedding dimension. The index of tokens runs from $0$ to $N$, and $0$ represents the \verb|[CLS]| token. The tokens $\mathbf{x}_{\mathrm{token}}$ are input into a sequence of attention layers.

In the MHA of each attention layer, three matrices are produced by transforming the input with three different trained linear projection layer: a query $\mathbf{Q}$, a key $\mathbf{K}$, and a value $\mathbf{V}$ $\in \realn^{(N+1)\times d}$. Then, the elements of $\mathbf{Q}$, $\mathbf{K}$, and $\mathbf{V}$ are divided among multiple heads indexed by $i$, where $\mathbf{K}_{i}\in \realn^{(N+1)\times d_{k}}, 
\quad i=1,2,\dots,h.$ $h$ is the number of the heads, and $d_{k}$ is the embedding dimension of $\mathbf{K}$. The same applies to $\mathbf{Q}$ and $\mathbf{V}$. The attention matrix of $i$ head, denoted by $A_{i}\in \realn^{(N+1)\times (N+1)}$ is defined as
\begin{align}
    \label{eq:attnmat}
    A_{i} = \mathrm{Softmax}\left(\frac{{\mathbf{Q}_{i}\mathbf{K}_{i}^\top}}{{\sqrt{d_k}}}\right).
\end{align}
Subsequently, $\mathrm{MHA}$ is calculated as
\begin{align}
    \label{eq:mha}
    &\mathrm{MHA} = \mathrm{Linear}\left(\left[\mathrm{head}_{1},\mathrm{head}_{2},\dots,\mathrm{head}_{h}\right]\right),\\
    \label{eq:headi}
    &\mathrm{head}_{i} = A_{i}\mathbf{V}_{i},
\end{align}
where $\mathrm{Linear}$ is a trained linear projection. To represent the feature of the image, we focus on the last attention layer. Let $\mathrm{x}_{*}$ and $\mathrm{MHA}_{*}$ be the input and $\mathrm{MHA}$ of the last attention layer respectively. Then, the output of the last attention layer $z_{*}$ is written as
\begin{align}
    \label{eq:lastattn}
    z_{*} &= y_{*} + \mathrm{FFN}\left(\mathrm{LN}\left(y_{*}\right)\right),
    ~\text{where}~y_{*} = \mathrm{x}_{*} + \mathrm{MHA}_{*},
\end{align}
where $\mathrm{LN}$ is layer normalization~\cite{ba2016layer} followed by a feed-forward network (FFN). The sums in \cref{eq:lastattn} correspond to residual connections~\cite{kaiming2016res}. Finally, the feature of the image $\mathbf{f}(\mathbf{x}_\mathrm{input})$ is obtained as the \verb|[CLS]| part of the layer-normalized output:
\begin{align}
    \label{eq:feature}
    \mathbf{f}\left(\mathbf{x}_\mathrm{input}\right) &= \left[\mathrm{LN}\left(z_{*}\right)\right]_{0}.
\end{align}

To perform image retrieval, we use $k$-nearest neighbors (NN) method with cosine similarity. Let $\mathcal{I}_{\mathrm{Q}}$ be the set of query images and $\mathcal{I}_{\mathrm{DB}}$ be the set of database images. Then, for a query image $\mathbf{x}_{\mathrm{Q}}\in\mathcal{I}_{\mathrm{Q}}$, the $k^\prime$th most similar image in the database $\mathbf{x}_{\mathrm{R}}(\mathbf{x}_{\mathrm{Q}},\mathcal{I}_{\mathrm{DB}},k^\prime)$ is retrieved as
\begin{align}
  \label{eq:cosine}
    \mathbf{x}_{\mathrm{R}}\left(\mathbf{x}_{\mathrm{Q}},\mathcal{I}_{\mathrm{DB}},k^\prime\right) = 
    \underset{\mathbf{x}_{\mathrm{DB}}\in\mathcal{I}_{\mathrm{DB}}}{\mathrm{argmax}_{k^\prime}}
    \left(\frac{\mathbf{f}\left(\mathbf{x}_{\mathrm{DB}}\right)^\top\mathbf{f}\left(\mathbf{x}_{\mathrm{Q}}\right)}{\norm{\mathbf{f}\left(\mathbf{x}_{\mathrm{DB}}\right)} \norm{\mathbf{f}\left(\mathbf{x}_{\mathrm{Q}}\right)}}\right),
\end{align}
where $\mathrm{argmax}_{k^\prime}(s)$ is a function that returns $\mathbf{x}_{\mathrm{DB}}$ with the $k^\prime$th largest cosine similarity. Consequently, the top-$k$ images can be retrieved by running \cref{eq:cosine} with $\forall k^\prime\leq k$.

\subsection{Prompt-Driven Attention Head Selection}
\label{subsec:phs}

Here, we present the algorithm for PHS. Our method involves the selection of $h_{\mathrm{on}}$ heads from a total of $h$ heads in the last attention layer, based on the visual prompt mask $\mathbf{M}_{\mathrm{input}}\in\left\{0, 1\right\}^{H\times W}$. The nonzero part of $\mathbf{M}_{\mathrm{input}}$ represents the region of interest (ROI) defined by the visual prompt. Initially, $\mathbf{M}_{\mathrm{input}}$ is converted to attention mask tokens denoted as $\mathbf{M}_{\mathrm{token}}\in \left\{0, 1\right\}^{N}$. Within this set of tokens, we identify patch tokens in the ROI, marked them as $T^{\mathrm{HS}} = \left\{t\in\left\{1,2,\dots,N\right\} \mid \mathbf{M}_{\mathrm{token},t} = 1\right\}$. Subsequently, we calculate the ROI attention $A_{i}^{\mathrm{HS}}$ for each attention head by summing up the attention values of $A_{i}$ in \cref{eq:attnmat}, but only for the tokens within $T^{\mathrm{HS}}$, defined by
\begin{align}
  \label{eq:phssum}
  A_{i}^{\mathrm{HS}} = \sum_{t\in T^{\mathrm{HS}}} A_{i,0,t}.
\end{align}
The straightforwardness of \cref{eq:phssum} in extracting ROI attentions facilitates the integration of diverse visual prompts, including point, box, and segmentation. This, in turn, streamlines the system design process in real-world implementations. Once the ROI attention for each head is computed, we set $h_{\mathrm{on}}$ heads with the highest $A_{i}^{\mathrm{HS}}$ values as our selected heads $i_\mathrm{on}\subseteq \left\{1,2,\dots,h\right\}$, which satisfies $|i_\mathrm{on}|=h_\mathrm{on}$. In our method, $h_{\mathrm{on}}$ serves as the only parameter. Our experiments in \cref{sec:experiments} demonstrate that $h_{\mathrm{on}}$ is sufficiently robust and can be set as a fixed value.

Subsequently, we replace the MHA in the last attention layer $\mathrm{MHA}_{*}$ with our approach, $\mathrm{MHA}_{*}^{\mathrm{HS}}$, defined by
\begin{align}
\label{eq:mhahs}
\mathrm{MHA}_{*}^{\mathrm{HS}} &= \mathrm{Linear}\left[\mathrm{head}_{*~1}^{\mathrm{HS}},\mathrm{head}_{*~2}^{\mathrm{HS}},\dots,\mathrm{head}_{*~h}^{\mathrm{HS}}\right], \\
\label{eq:mhahsih}
\mathrm{head}_{*~i}^{\mathrm{HS}} &= 
    \begin{cases}
      \mathrm{head}_{*~i} \times h/h_\mathrm{on} & \text{if $i \in i_\mathrm{on}$,} \\
        0 & \text{if $i \notin i_\mathrm{on}$.}
    \end{cases}    
\end{align}
Our head selection strategy is inspired by Ref.~\cite{nicolicioiu2023learning}, where head selection is performed prior to the output linear projection layer of MHA and $\mathrm{head}_{*~i}$ is multiplied by a scaling factor of $h/h_\mathrm{on}$. Scaling factor helps to preserve the overall attention intensity in MHA. Following the MHA, our subsequent process adheres to the standard retrieval pipeline by applying \cref{eq:mhahs} to \cref{eq:lastattn,eq:feature,eq:cosine}. Importantly, our method does not necessitate any model fine-tuning. The head selection operation is performed to a trained ViT model during inference, dynamically matching the attention head to the user-defined visual prompt, as shown in \cref{fig:overviewmethod}. 

One intriguing aspect of our method is its capability to simultaneously apply head selection to both query and database images using only the query visual prompt. In contrast, standard prompt-based methods such as FA~\cite{falip} typically can only be applied on the query image, with a fixed retrieval database. Our approach offers two distinct modes of operation: (1) Query-Only PHS and (2) Query-DB PHS. The retrieval process of Query-Only PHS mode is compatible with standard prompt-based methods, where PHS is performed solely on the query image. In contrast, Query-DB PHS mode extends the head selection process to the images in the retrieval database, dynamically adapting it for each query. Specifically, this mode modifies each feature in the retrieval database by performing head selection with the same attention heads selected by using the query. By doing so, Query-DB PHS intuitively enhances the feature space of both the query and retrieval database with a query prompt, improving performance in certain scenarios. We mainly report the results of Query-Only PHS as our method in this paper for its compatible retrieval process.

\section{Experiments}
\label{sec:experiments}

\subsection{Experimental Settings}
\label{subsec:expsettings}

\begin{figure}[h]
  \centering
  \includegraphics[width=1.0\linewidth]{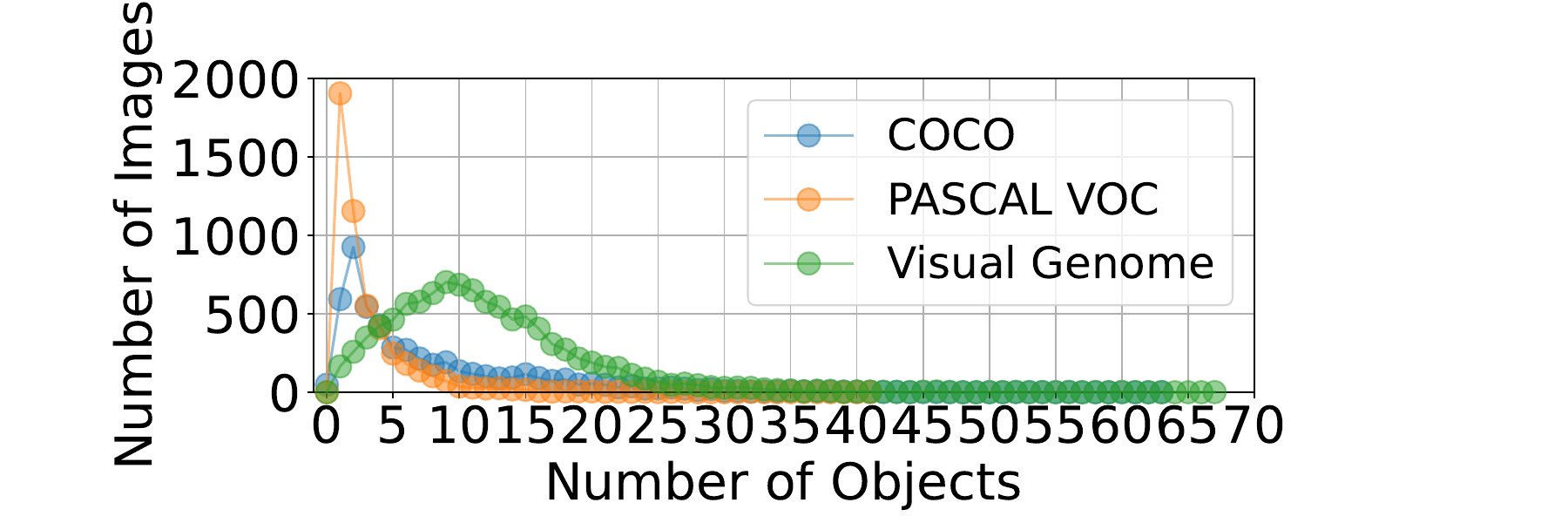}
  \caption{Histogram of number of objects per query in datasets.}
  \label{fig:hist}
\end{figure}

We evaluate the image retrieval performance of our method on three multi-object-based image datasets: COCO~\cite{lin2014coco}, PASCAL VOC~\cite{everingham2010pascalvoc} and Visual Genome~\cite{visualgenome}. We choose these datasets as they contain images with multiple objects and intricate background in general, a fitting scenario to FOIR task. COCO dataset contains 80 object categories, and we use val2017 for query images $\mathcal{I}_{\mathrm{Q}}$ and train2017 for database images $\mathcal{I}_{\mathrm{DB}}$, where $|\mathcal{I}_{\mathrm{Q}}| = 5000$ and $|\mathcal{I}_{\mathrm{DB}}| = 118287$. PASCAL VOC dataset contains 20 object categories, and we use test2007 for $\mathcal{I}_{\mathrm{Q}}$ and trainval2007$+$2012 for $\mathcal{I}_{\mathrm{DB}}$, where $|\mathcal{I}_{\mathrm{Q}}| = 4952$ and $|\mathcal{I}_{\mathrm{DB}}| = 16551$. Visual Genome dataset contains 33877 object categories. We select the most frequent 100 categories taking into account object aliases. We use the test and training sets in Ref.~\cite{Chen2019LearningSG} as $\mathcal{I}_{\mathrm{Q}}$ and $\mathcal{I}_{\mathrm{DB}}$ excluding images without categorized objects from $\mathcal{I}_{\mathrm{Q}}$, then $|\mathcal{I}_{\mathrm{Q}}| = 9880$ and $|\mathcal{I}_{\mathrm{DB}}| = 82904$. The input images of all datasets are resized to $224 \times 224$ pixels. \cref{fig:hist} shows histogram illustrating the distribution of the number of objects per query across the datasets utilized. Notably, COCO and Visual Genome datasets predominantly feature multi-object queries, which fit the definition of FOIR task, while 38\% of the queries in PASCAL VOC dataset are single-object queries.

For pretrained ViT models, we utilize the publicly available DINOv2 models with four different sizes: \textit{small, base, large, giant}. We specifically use the models trained with register tokens~\cite{darcet2023vision} to address any unwanted artifacts in attention patches. Note that we do not perform additional training on these models. The number of available heads ($h$) varies depending on the model size: 6 for \textit{small}, 12 for \textit{base}, 16 for \textit{large}, and 24 for \textit{giant}. For our experiments, we set the number of selected heads ($h_{\mathrm{on}}$) to 5, which we found to be generally robust across all model types unless stated otherwise. Although the optimum $h_{\mathrm{on}}$ can be slightly model-dependent due to the scaling of the number of heads, detailed parameter tuning is not necessary based on our observations.

To quantitatively compare our proposed method, we evaluate it against two baselines. The first baseline, called \textit{CBIR}, is a standard CBIR implementation using DINOv2 models. The second baseline, referred to as \textit{Mask}, is a prompt-driven image alteration method where the region outside the region-of-interest is masked before CBIR is performed using DINOv2 models. These baselines represent strong non-prompt-based and prompt-based approaches respectively, as DINOv2 is already a powerful model for image retrieval tasks. Note that we exclude the crop-based technique from our main comparison as it significantly alters the image and object size, which changes the query format and task characteristics. Additionally, we compare our method to Foveal Attention (\textit{FA})~\cite{falip}, a state-of-the-art non-image-alteration visual prompting technique that is most relevant to our study. For \textit{FA}, we follow the original work and implement it in the last 4 layers of the ViT model. To ensure a fair comparison, we use box prompts for all experiments unless stated otherwise, as \textit{FA} is designed for bounding box prompts. We use the box labels in each dataset to simulate visual prompt inputs. Each box label represents a single query. It's worth noting that our method is flexible and supports various types of prompt inputs, such as point, box, and segmentation prompts. We demonstrate their superior performance in ~\cref{subsec:prompttype}. Additional analysis can be found in the supplementary material.

In our experiments, we assess the retrieval performance of our method using two metrics: Mean Precision at $k$ ($\mathrm{MP@}k$) and Mean Average Precision at $k$ ($\mathrm{MAP@}k$). $\mathrm{MP@}k$ is used to evaluate the balanced retrieval results for a selected value of $k$, which in our case is $k=100$. On the other hand, $\mathrm{MAP@}k$ is employed to evaluate the higher-ranking retrieval performance of our method.

\subsection{Focus-Oriented Image Retrieval Results}
\label{subsec:mainresult}

\begin{table}[h]
  \begin{center}
  \resizebox{1.0\columnwidth}{!}{
  \begin{tabular}{|c|c|c@{ }@{ }c@{ }@{ }c@{ }@{ }c|c@{ }@{ }c@{ }@{ }c@{ }@{ }c|}
  \hline
  Dataset & Model & \multicolumn{4}{|c|}{$\mathrm{MP@}100~(\mathrm{\%})$} & \multicolumn{4}{|c|}{$\mathrm{MAP@}100~(\mathrm{\%})$}\\
          & Size  & CBIR\cite{darcet2023vision} & Mask & FA\cite{falip} & Ours & CBIR\cite{darcet2023vision} & Mask & FA\cite{falip} & Ours\\
  \hline
  \multirow{4}{*}{COCO}
        &small & 54.8 & 35.1       & {\bf 55.3} & 54.9       & 59.0 & 38.0       & {\bf 59.6} & 59.2       \\
        &base  & 57.4 & 43.2       & 57.9       & {\bf 60.6} & 61.5 & 45.9       & 62.0       & {\bf 64.1} \\
        &large & 58.4 & 47.5       & 58.8       & {\bf 61.3} & 62.4 & 50.0       & 62.8       & {\bf 65.1} \\
        &giant & 58.5 & 50.4       & 58.8       & {\bf 60.7} & 62.7 & 52.9       & 62.9       & {\bf 64.8} \\
  \hline
  \multirow{4}{*}{\shortstack{PASCAL\\ VOC}}
        &small & 77.2 & 72.7       & {\bf 77.8} & 77.4       & 79.8 & 75.0       & {\bf 80.4} & 80.2       \\
        &base  & 78.6 & 79.0       & 79.0       & {\bf 80.9} & 81.1 & 81.4       & 81.5       & {\bf 83.7} \\
        &large & 77.8 & {\bf 80.4} & 78.1       & 80.3       & 81.0 & {\bf 83.6} & 81.3       & {\bf 83.6} \\
        &giant & 78.3 & {\bf 82.4} & 78.5       & 79.6       & 81.1 & {\bf 85.3} & 81.3       & 82.5       \\ 
  \hline
  \multirow{4}{*}{\shortstack{Visual\\ Genome}}
        &small & 30.1 & 20.5       & {\bf 30.2} & 30.1       & 34.3 & 24.0       & {\bf 34.5} & 34.3       \\
        &base  & 29.6 & 23.9       & 29.8       & {\bf 31.4} & 34.1 & 27.5       & 34.2       & {\bf 35.6} \\
        &large & 29.1 & 24.3       & 29.1       & {\bf 30.2} & 33.7 & 28.5       & 33.8       & {\bf 34.7} \\
        &giant & 29.1 & 26.1       & 29.2       & {\bf 29.9} & 33.8 & 30.2       & 33.8       & {\bf 34.5} \\ 
  \hline
  \end{tabular}
  }
  \end{center}
  \caption{FOIR results with DINOv2 models. \textit{CBIR} denotes a standard CBIR implementation. \textit{Mask} denotes a mask based prompt-guided method. \textit{FA} denotes Foveal Attention, a state-of-the-art visual prompting technique. \textit{Ours} denotes our method. Our method outperforms the baselines and \textit{FA} in most cases (Model: DINOv2).}
  \label{tab:mainresult}
\end{table}

The results of the proposed method, retrieval with \textit{FA} employed, and the baselines are summarized in \cref{tab:mainresult}. It can be observed that our method generally improves the performance for both $\mathrm{MP@}100$ and $\mathrm{MAP@}100$ evaluations. Improvements in both $\mathrm{MP@}100$ and $\mathrm{MAP@}100$ indicates that our method improves not only the top-$100$ retrieval accuracy but also higher ranked images. Substantial improvements are particularly observed when comparing our method to the baselines of \textit{CBIR} and \textit{Mask}. The absence of visual prompt in \textit{CBIR} limits its accuracies, while \textit{Mask's} reduction of visual context leads to inconsistent performance in the FOIR task across multiple datasets. In \cref{subsec:effectiveness}, we analyse the higher performance observed for \textit{Mask} in certain cases of PASCAL VOC.

Comparing our method to utilizing the recent visual prompting technique \textit{FA}, we achieve substantial enhancements in accuracies, except for the \textit{small} model. We speculate that the number of heads in the ViT models plays a role in the effectiveness of our method. The \textit{small} model, with only 6 heads, may lack sufficient semantic differentiation to effectively implement a head selection algorithm. However, as we scale up to larger models, the increased number of heads enables our method to be more effective, which aligns with the concept of our approach. We also observe that \textit{FA} consistently performs effectively across all model sizes and datasets, demonstrating the robustness of its attention manipulation technique.

\subsection{Different Pretraining Paradigm Generalization}
\label{subsec:clip}
Here, we examine the ability of our method to generalize to models with different pretraining approach. We assess the performance with CLIP~\cite{clip,openclip} models, which are pretrained using a large-scale image-text weakly supervised strategy, in \cref{tab:clipresult}. CLIP offers a distinct perspective compared to DINOv2, which is pretrained using SSL strategy. We evaluate pretrained ViT-B/16 and ViT-L/14 CLIP vision models in our experiments. The results demonstrate that our method consistently outperforms the baselines and \textit{FA} in all scenarios, indicating its strong generalization capability. We speculate that CLIP learned strong user-oriented semantic attentions through its weakly supervised pretraining strategy, leading to its exceptional compatibitility with our method.

\begin{table}[h]
  \begin{center}
  \resizebox{1.0\columnwidth}{!}{
  \begin{tabular}{|c|c|c@{ }@{ }c@{ }@{ }c@{ }@{ }c|c@{ }@{ }c@{ }@{ }c@{ }@{ }c|}
  \hline
  Dataset & Model & \multicolumn{4}{|c|}{$\mathrm{MP@}100~(\mathrm{\%})$} & \multicolumn{4}{|c|}{$\mathrm{MAP@}100~(\mathrm{\%})$}\\
          & Size  & CBIR\cite{clip} & Mask & FA\cite{falip} & Ours & CBIR\cite{clip} & Mask & FA\cite{falip} & Ours\\
  \hline
  \multirow{2}{*}{COCO}
            &base  & 52.6 & 27.4 & 53.3 & {\bf 55.7} & 57.9 & 30.7 & 58.5 & {\bf 60.2}\\
            &large & 54.7 & 33.4 & 55.2 & {\bf 58.0} & 59.7 & 37.3 & 60.2 & {\bf 62.6}\\
  \hline
  \multirow{2}{*}{\shortstack{PASCAL\\ VOC}}
            &base  & 71.5 & 60.9 & 72.2 & {\bf 73.8} & 76.0 & 65.5 & 76.6 & {\bf 77.9}\\
            &large & 71.4 & 63.8 & 72.0 & {\bf 74.2} & 76.0 & 69.6 & 76.5 & {\bf 78.4}\\
  \hline
  \multirow{2}{*}{\shortstack{Visual\\ Genome}}
            &base  & 29.3 & 15.3 & 29.5 & {\bf 30.1} & 34.1 & 18.7 & 34.3 & {\bf 34.8}\\
            &large & 28.9 & 17.0 & 29.1 & {\bf 30.2} & 33.8 & 21.5 & 33.9 & {\bf 34.9}\\
  \hline
  \end{tabular}
  }
  \end{center}
  \caption{FOIR results with CLIP models. Our method outperforms the baselines and \textit{FA} in all cases, demonstrating the model generalization capability (Model: CLIP).}
  \label{tab:clipresult}
\end{table}

\subsection{Analysis on Number of Objects}
\label{subsec:effectiveness}

\begin{figure}[h]
  \begin{minipage}{1.0\linewidth}
    \centering
    \includegraphics[width=1.0\linewidth]{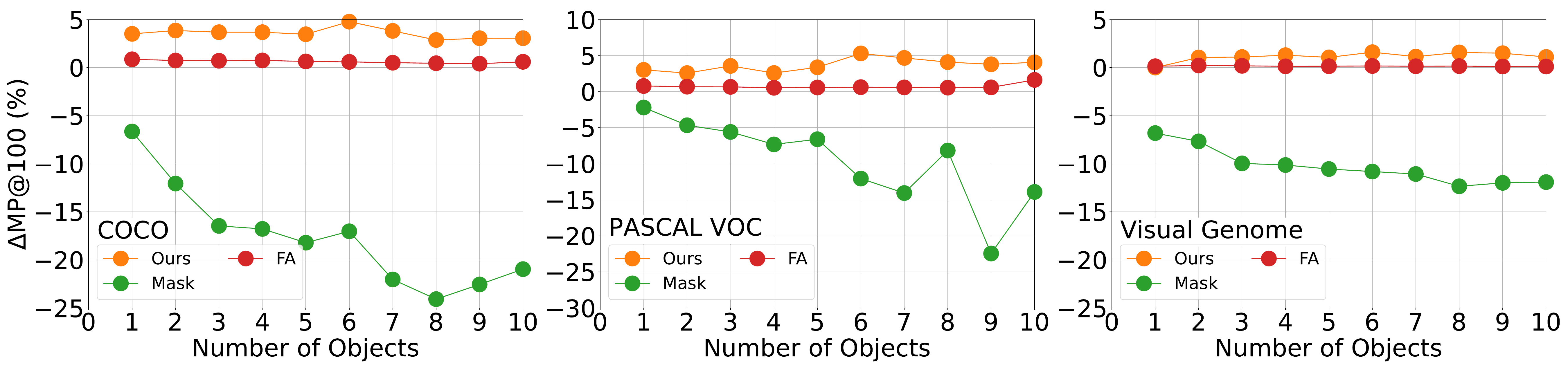}
    \caption{Relative performance to CBIR (Model: CLIP \textit{large}).}
    \label{fig:objects-clip}
  \end{minipage}
  \begin{minipage}{1.0\linewidth}
    \centering
    \includegraphics[width=1.0\linewidth]{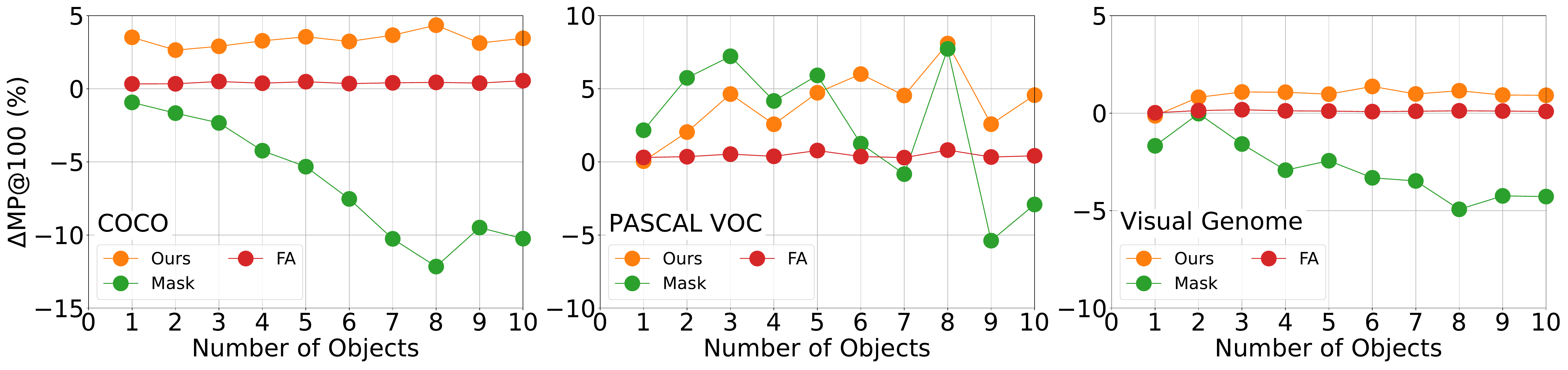}
    \caption{Relative performance to CBIR (Model: DINOv2 \textit{large}).}
    \label{fig:objects-dinov2}
  \end{minipage}
\end{figure}

\cref{fig:objects-clip} and \cref{fig:objects-dinov2} illustrate the relative performance of methods with CLIP and DINOv2 models with respect to CBIR, across varying numbers of objects in the query. Our analyses reveal that our method performs particularly well for multi-object queries (two or more objects), especially within the Visual Genome and PASCAL VOC datasets, which aligns with the anticipated outcomes of our approach. In contrast, the \textit{Mask} method demonstrates instability when handling a larger number of objects. Notably, for DINOv2 model, our method achieves $\mathrm{MP@}100$ of 77.9\%, surpassing the \textit{Mask} method's 77.2\% when focusing solely on the multi-object queries subset of the PASCAL VOC. The strong performance of \textit{Mask} method for DINOv2 model on single-object queries for PASCAL VOC (\cref{fig:objects-dinov2}) significantly contributes to its overall superior accuracy in comparison to our method (\cref{tab:mainresult}), as 38\% of the queries in PASCAL VOC are single-object queries. Nevertheless, our approach demonstrates a distinct advantage as the number of objects increases.

\subsection{Robustness on Various Visual Prompt Types}
\label{subsec:prompttype}

\begin{table}[h]
  \begin{center}
  \resizebox{1.0\columnwidth}{!}{
  \begin{tabular}{|c|c|ccc|ccc|}
  \hline
  Model & CBIR                    & \multicolumn{3}{|c|}{Mask} & \multicolumn{3}{|c|}{Ours}\\
  Size  & \cite{darcet2023vision} & Point & Box & Segment & Point & Box & Segment\\
  \hline
  small & 54.8 & 2.4 & 35.1 & 30.5 & 54.7 & 54.9 & 55.1\\
  base  & 57.4 & 3.3 & 43.2 & 37.3 & 59.7 & 60.6 & 61.3\\
  large & 58.4 & 3.7 & 47.5 & 40.6 & 61.2 & 61.3 & 61.9\\
  giant & 58.5 & 5.6 & 50.4 & 44.4 & 60.6 & 60.7 & 61.0\\
  \hline
  \end{tabular}
  }
  \end{center}
  \caption{FOIR results with various prompt types (Dataset: COCO, Model: DINOv2, Metric: $\mathrm{MP@}100~(\mathrm{\%})$).}
  \label{tab:promptresult}
\end{table}

Our method incorporates a prompt-based attention head matching aspect, allowing it to effectively handle various types of visual prompts such as points, boxes, or segmentations. Unlike standard prompt-based methods like \textit{Mask}, which heavily rely on a strict region of interest, our method is robust across all forms of visual prompts, as demonstrated in \cref{tab:promptresult}. For instance, a simple point (click) prompt with a fixed window size is sufficient for head selection, and even imperfect prompts with arbitrary shapes are expected to yield satisfactory results. In contrast, the \textit{Mask} approach's performance noticeably declines when using segment and point prompts. Note that, in the case of point prompt, the user input may vary and not accurately represent the center of the object. To address this, we perform experiments on every possible patch window position within the segmentation mask and calculate the aggregated accuracies. It is important to mention that an evaluation of \textit{FA} is not conducted due to its algorithm's incompatibility with segment and point prompts.

\subsection{Visual Analysis with Attention Map}
\label{subsec:visualatt}
\begin{figure}[h]
  \centering
  \includegraphics[width=0.9\linewidth]{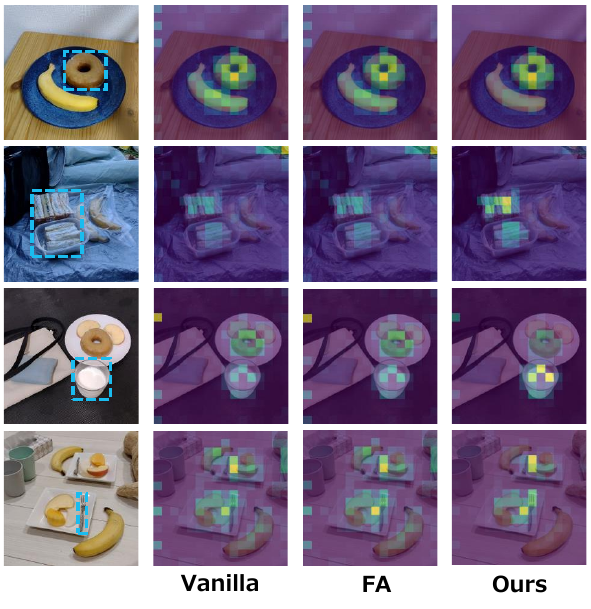}
  \caption{The visualization of attention maps demonstrates that our method performs more intuitively than \textit{Vanilla} ViT and \textit{FA}. Best viewed in color (Model: DINOv2 \textit{giant}).}
  \label{fig:visualize}
\end{figure}

Here, we present the results of our visualization analysis on attention maps generated in the final layer of the ViT model, after incorporating our proposed method. As depicted in \cref{fig:visualize}, our method demonstrates superior intuition in terms of enhanced focus and noise reduction when comparing to \textit{Vanilla} ViT (used in \textit{CBIR}) and \textit{FA}. In contrast, \textit{FA} typically generates attention maps that are comparable to those produced by \textit{Vanilla} ViT, albeit with slightly more concentrated ROI attentions. Note that our approach preserves potentially valuable surrounding visual context, which plays a crucial role in reflecting user perception.

\subsection{Image Alteration \& Prompt Noise Analysis}
\label{subsec:formatandnoise}
In this section, we conduct image alteration study on image retrieval using a crop-based visual prompt to emulate the IRQ query format. It is important to note that this task setting is different from FOIR, which employs the WIQ query format. Here, we investigate their exposure to prompt error by adding noise to the box prompt. Users typically do not generate perfect prompts, necessitating the ability of a prompt-driven method to tolerate some level of noise. To simulate this, we introduce noise into the prompts by randomly shifting and resizing the visual prompts by roughly 7.6\% of image width and height in average. The results, as depicted in \cref{tab:formatandnoise}, demonstrate that our method's accuracies (in WIQ query format) remain consistently stable even in the presence of prompt noise. This suggests that our method effectively handles imperfect prompts due to its perception matching mechanism. Conversely, the performance of \textit{Crop} query deteriorates when exposed to prompt noise, highlighting the limitations of image alteration technique.

\begin{table}[h]
  \begin{center}
  \resizebox{1.0\linewidth}{!}{
  \begin{tabular}{|c|cccc|cccc|}
  \hline
  Model & \multicolumn{4}{|c|}{DINOv2} & \multicolumn{4}{|c|}{CLIP}\\
  Size  & Crop & Crop-N & Ours & Ours-N & Crop & Crop-N & Ours & Ours-N\\
  \hline
  small & 60.0 & 44.9 & 54.9 & 54.8 & -    & -    & -    & -   \\
  base  & 66.7 & 49.6 & 60.6 & 59.6 & 45.2 & 36.2 & 55.7 & 55.2\\
  large & 67.3 & 50.6 & 61.3 & 60.6 & 52.3 & 40.7 & 58.0 & 57.6\\
  giant & 68.7 & 51.6 & 60.7 & 60.4 & -    & -    & -    & -   \\
  \hline
  \end{tabular}
  }
  \end{center}
  \caption{Image retrieval results with image alteration and noise. Method names end with \textit{-N} represent noisy prompt (Dataset: COCO, Metric: $\mathrm{MP@}100~(\mathrm{\%})$).}
  \label{tab:formatandnoise}
\end{table}

\subsection{Qualitative Case Study on Visual Context}
\label{subsec:visualcontext}

\begin{figure}[h]
  \centering
  \includegraphics[width=1.0\linewidth]{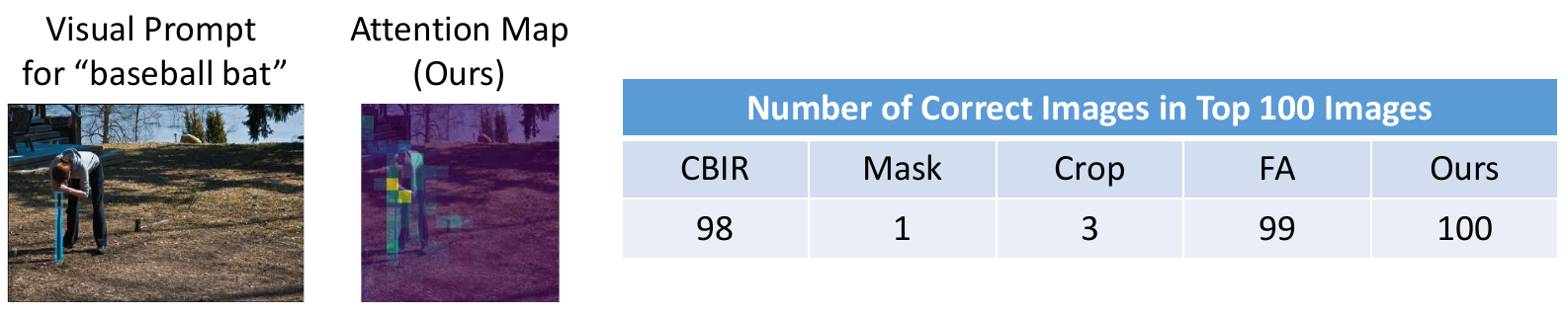}
  \caption[]{Qualitative case study on visual context consideration. \footnotemark }
  \label{fig:visualcontext}
\end{figure}
\footnotetext{license for the image is provided in supplementary material}

In \cref{fig:visualcontext}, we present a qualitative case study on a query image from COCO dataset with a label of \textit{baseball bat}. In this case, simply cropping or masking the object leads to significant deterioration in retrieval performance, while our method further increases accuracy, demonstrating the necessity of surrounding visual context.

\subsection{Analysis on Number of Selected Heads}
\label{subsec:numheads} 
\begin{figure}[h]
  \centering
  \includegraphics[width=1.0\linewidth]{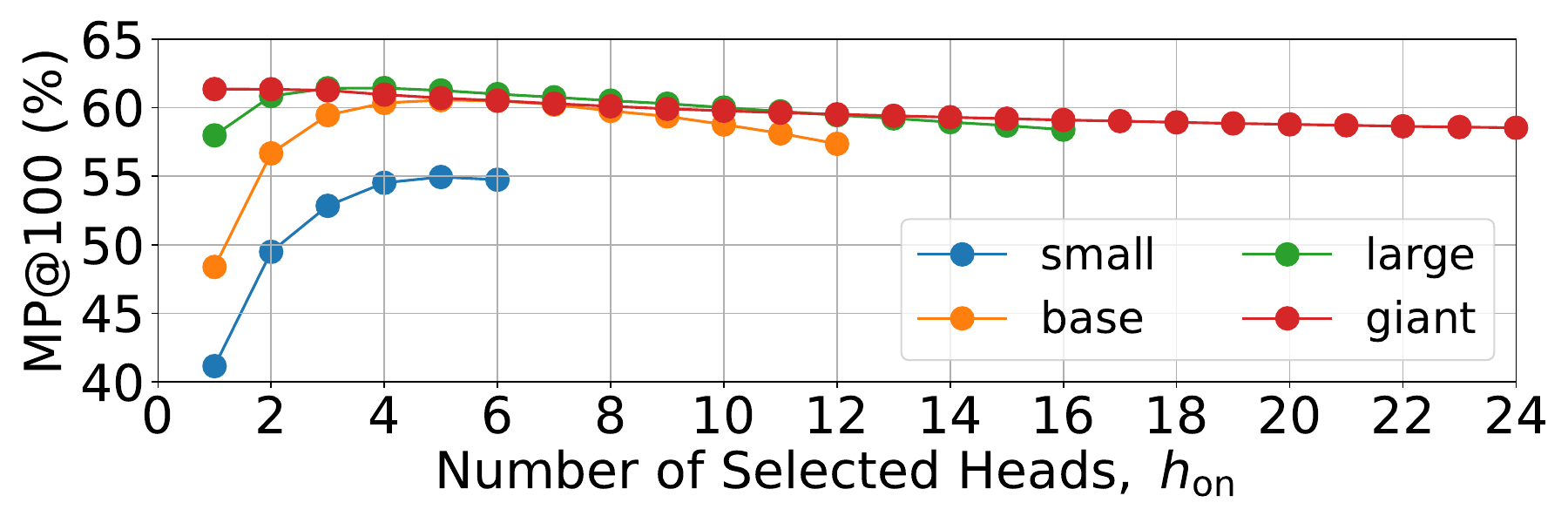}
  \caption{FOIR results with various number of selected heads (Dataset: COCO, Model: DINOv2).}
  \label{fig:heads}
\end{figure}

Here, we investigate the parameter of the number of selected heads by performing a parameter scan for $h_{\mathrm{on}}$. In \cref{fig:heads}, we vary the number of heads across different model sizes. We observe that the optimal number of heads is around 5 for \textit{large, base,} and \textit{small} models. Based on this observation, we selected 5 as our common parameter. Interestingly, \textit{giant} model exhibits the highest accuracy with only 1 selected head. We believe that the \textit{giant} model, with its sufficiently deep and large architecture, has learned more precise semantic information across the attention heads.

\section{Conclusion}
\label{sec:conclusion}
We proposed Prompt-guided attention Head Selection to leverage the head-wise potential of the multi-head attention mechanism in ViT for image retrieval. We setup the Focus-Oriented Image Retrieval task to simulate real-world scenarios with complex images and user interest in retrieving images with specific object. Our method matched user perceptions to attention heads, bridging the gap between human and model visual understanding. PHS does not require model re-training or image alteration, ensuring no undesirable consequences of image editing. Experiments showed that PHS improves performance on multiple datasets, enhancing ViT model in the FOIR task.

{
    \small
    \bibliographystyle{ieeenat_fullname}
    \bibliography{mainbib}
}

\clearpage

\newcommand\beginsupplement{%
         \setcounter{equation}{0}
         \renewcommand{\theequation}{A\arabic{equation}}
         \setcounter{figure}{0}
         \renewcommand{\thefigure}{A\arabic{figure}}
         \setcounter{table}{0}
         \renewcommand{\thetable}{A\arabic{table}}
         \setcounter{section}{0}
         \renewcommand{\thesection}{A\arabic{section}}
}
\beginsupplement

\section*{Supplementary Material}

\section{Extended Analysis}

\subsection{Augmenting Current Method with PHS}
\label{subsec:faours}
In this study, we examine the performance of augmenting the \textit{FA} method with our proposed approach. The \textit{FA} method involves an attention additive operation, while our method involves an attention head selection operation. These two methods can be implemented simultaneously without any conflicts. The experimental results, presented in \cref{tab:facombineresult}, clearly indicate a complementary relationship between \textit{FA} and our method. Notably, the combined approach, denoted as \textit{FA+Ours}, achieves the highest accuracies across all experimental conditions.

\begin{table}[h]
  \begin{center}
  \resizebox{1.0\columnwidth}{!}{
  \begin{tabular}{|c|c|cccc|}
  \hline
  Dataset & Model & \multicolumn{4}{|c|}{Method}\\
          & Size  & CBIR\cite{darcet2023vision} & FA\cite{falip} & Ours & FA+Ours\\
  \hline
  \multirow{4}{*}{COCO}
            &small & 54.8 & 55.3       & 54.9       & {\bf 55.5}\\
            &base  & 57.4 & 57.9       & 60.6       & {\bf 61.1}\\
            &large & 58.4 & 58.8       & 61.3       & {\bf 61.6}\\
            &giant & 58.5 & 58.8       & 60.7       & {\bf 61.0}\\
  \hline
  \multirow{4}{*}{\shortstack{PASCAL\\ VOC}} 
            &small & 77.2 & 77.8       & 77.4       & {\bf 77.9}\\  
            &base  & 78.6 & 79.0       & 80.9       & {\bf 81.2}\\
            &large & 77.8 & 78.1       & 80.3       & {\bf 80.5}\\
            &giant & 78.3 & 78.5       & 79.6       & {\bf 79.8}\\
  \hline
  \multirow{4}{*}{\shortstack{Visual\\ Genome}}
            &small & 30.1 & {\bf 30.2} & 30.1       & {\bf 30.2}\\  
            &base  & 29.6 & 29.8       & 31.4       & {\bf 31.5}\\
            &large & 29.1 & 29.1       & 30.2       & {\bf 30.3}\\
            &giant & 29.1 & 29.2       & {\bf 29.9} & {\bf 29.9}\\
  \hline
  \end{tabular}
  }
  \end{center}
  \caption{FOIR results when augmenting \textit{FA} with our method (Model: DINOv2, Metric: $\mathrm{MP@}100~(\mathrm{\%})$).}
  \label{tab:facombineresult}
\end{table}

\subsection{Method Variations \& Parameter Analysis}
\label{subsec:variations}

\begin{table}[h]
  \begin{center}
  \resizebox{1.0\linewidth}{!}{
  \begin{tabular}{|c|c|ccc|ccc|}
  \hline
  Dataset & Model & \multicolumn{3}{|c|}{DINOv2} & \multicolumn{3}{|c|}{CLIP}\\
          & Size  & FA\cite{falip} & Ours(QO) & Ours(QD) & FA\cite{falip} & Ours(QO) & Ours(QD)\\
  \hline
  \multirow{4}{*}{COCO}
        &small & {\bf 55.3} & 54.9       & 55.2       & -    & -          & -         \\
        &base  & 57.9       & {\bf 60.6} & 60.5       & 53.3 & 55.7       & {\bf 55.8}\\
        &large & 58.8       & {\bf 61.3} & 60.8       & 55.2 & 58.0       & {\bf 58.4}\\
        &giant & 58.8       & 60.7       & {\bf 61.4} & -    & -          & -         \\
  \hline
  \multirow{4}{*}{\shortstack{PASCAL\\ VOC}}   
        &small & {\bf 77.8} & 77.4       & 77.3       & -    & -          & -         \\
        &base  & 79.0       & {\bf 80.9} & 80.6       & 72.2 & {\bf 73.8} & 73.5      \\
        &large & 78.1       & {\bf 80.3} & 79.8       & 72.0 & {\bf 74.2} & 73.7      \\
        &giant & 78.5       & 79.6       & {\bf 79.9} & -    & -          & -         \\ 
  \hline
  \multirow{4}{*}{\shortstack{Visual\\ Genome}}  
        &small & {\bf 30.2} & 30.1       & {\bf 30.2} & -    & -          & -         \\
        &base  & 29.8       & {\bf 31.4} & 31.3       & 29.5 & {\bf 30.1} & 29.8      \\
        &large & 29.1       & 30.2       & {\bf 30.4} & 29.1 & {\bf 30.2} & 29.9      \\
        &giant & 29.2       & 29.9       & {\bf 30.0} & -    & -          & -         \\ 
  \hline
  \end{tabular}
  }
  \end{center}
  \caption{FOIR results of method variations. \textit{Ours(QO)} denotes our method with Query-Only PHS. \textit{Ours(QD)} denotes our method with Query-DB PHS (Metric: $\mathrm{MP@}100~(\mathrm{\%})$).}
  \label{tab:qoqdresult}
\end{table}

\begin{figure}[h]
  \centering
  \includegraphics[width=1.0\linewidth]{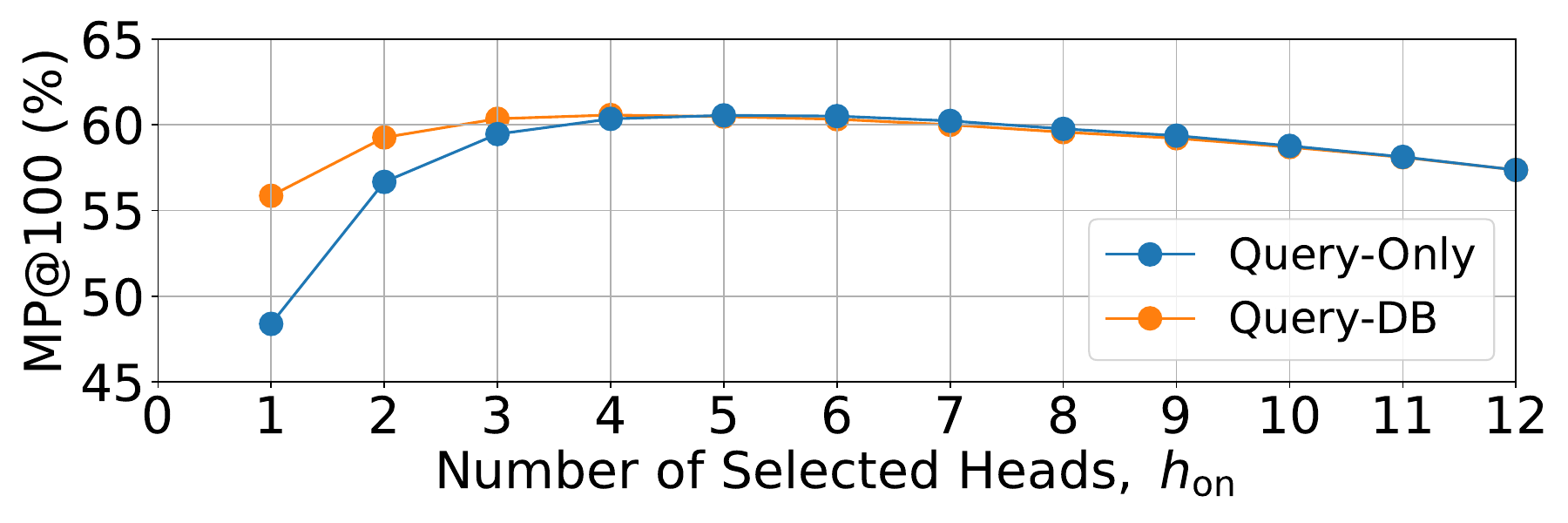}
  \caption{FOIR results with various number of selected heads (Dataset: COCO, Model: DINOv2 \textit{base}).}
  \label{fig:numheadsresult}
\end{figure}

Our approach offers two distinct modes of operation: (1) Query-Only PHS and (2) Query-DB PHS. The retrieval process of Query-Only PHS mode is compatible with standard prompt-based methods, where PHS is performed solely on the query image. In contrast, Query-DB PHS mode extends the head selection process to the images in the retrieval database, dynamically adapting it for each query. Specifically, this mode modifies each feature in the retrieval database by performing head selection with the same attention heads selected by using the query. By doing so, Query-DB PHS intuitively enhances the feature space of both the query and retrieval database with a query prompt, improving performance in certain scenarios. We mainly report the results of Query-Only PHS as our method in the main paper for its compatible retrieval process. Here, we report additional results for comparing Query-Only PHS and Query-DB PHS.

In our method variations, both Query-Only PHS (\textit{QO}) and Query-DB PHS (\textit{QD}) perform similarly and outperform \textit{FA} generally, as shown in \cref{tab:qoqdresult}. This indicates that applying our method to query only is sufficient to improve overall performance, while \textit{QD} shows its advantage in certain conditions, highlighting the effectiveness of database-side PHS. The advantage of \textit{QD} can be observed by investigating the parameter of the number of selected heads, $h_{\mathrm{on}}$. We perform a parameter scan for $h_{\mathrm{on}}$. The results in \cref{fig:numheadsresult} indicate that while both variations achieve similar performance when $h_{\mathrm{on}}$ is set to 5, \textit{QD} demonstrates superior robustness in the selection of $h_{\mathrm{on}}$. This enhancement can be attributed to its retrieval database side PHS component.

Modifying the retrieval database in \textit{QD} incurs higher computational costs. However, the head selection process occurs on the last layer, allowing for caching of query, key, and value features before the attention module. This means that only calculations in half of the last layer are needed, which can be efficiently achieved through GPU parallel processing. Additionally, since $\mathrm{LN}$ and $\mathrm{FFN}$ operations in Eq. (4) or Eq. (5) of the main paper are applied independently to each token, only the \verb|[CLS]| token needs to be extracted and calculated, further reducing computational requirements.

\subsection{Extended Analysis on Number of Objects}
\label{subsec:objects}

\begin{figure}[h]
  \begin{subfigure}{1.0\linewidth}
    \centering
    \includegraphics[width=1.0\linewidth]{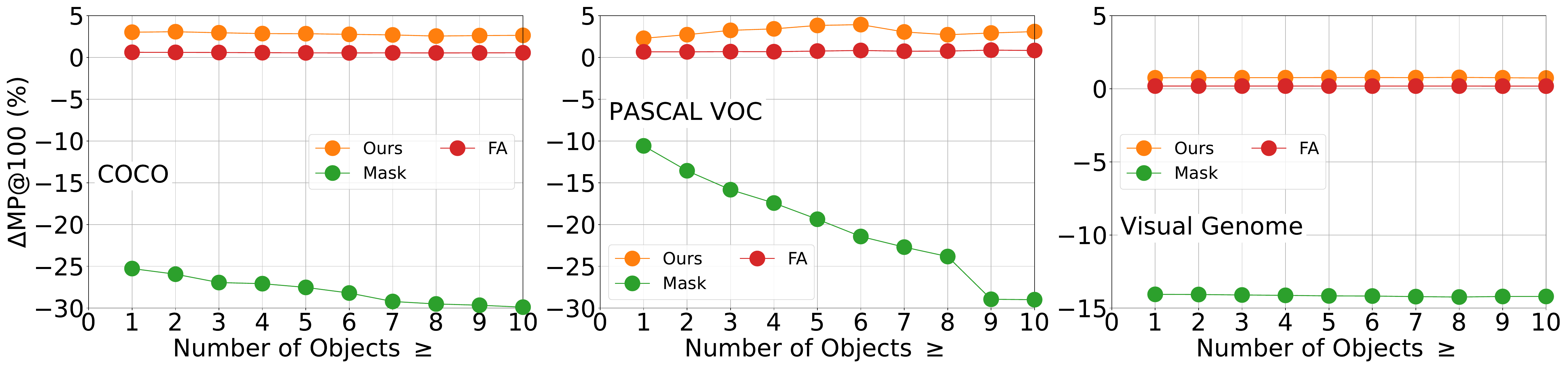}
    \caption{CLIP \textit{base}}
    \label{fig:objects-clip-base}
    \vspace{0.3cm}
  \end{subfigure}
  \begin{subfigure}{1.0\linewidth}
    \centering
    \includegraphics[width=1.0\linewidth]{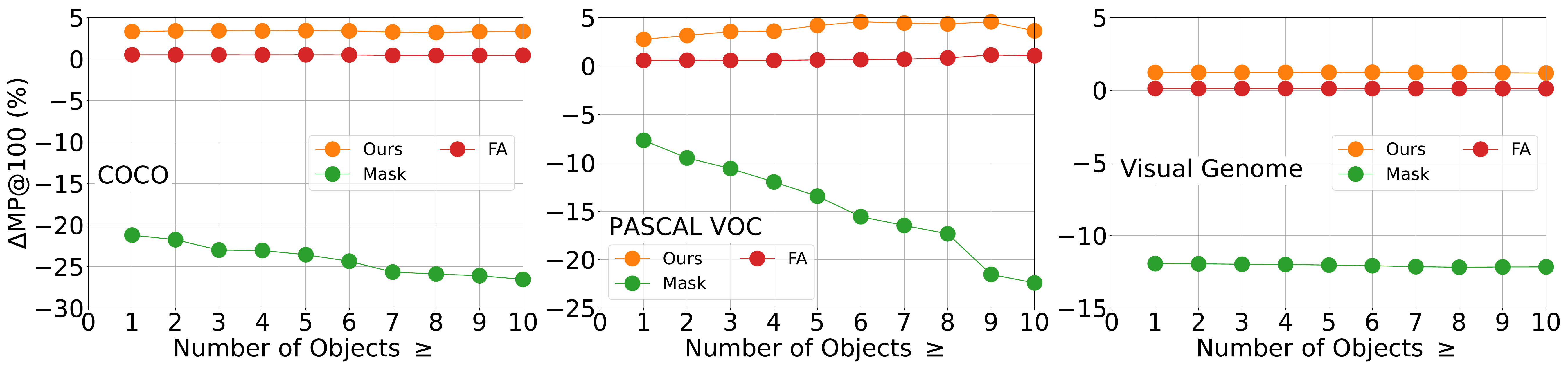}
    \caption{CLIP \textit{large}}
    \label{fig:objects-clip-large}
    \vspace{0.5cm}
  \end{subfigure}
  \begin{subfigure}{1.0\linewidth}
    \centering
    \includegraphics[width=1.0\linewidth]{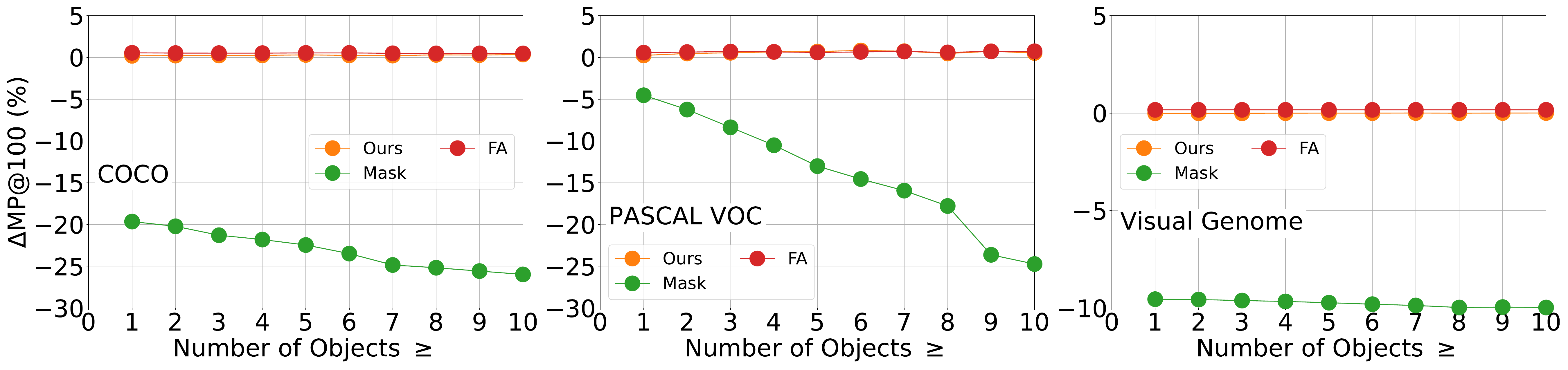}
    \caption{DINOv2 \textit{small}}
    \label{fig:objects-dinov2-small}
    \vspace{0.3cm}
  \end{subfigure}
  \begin{subfigure}{1.0\linewidth}
    \centering
    \includegraphics[width=1.0\linewidth]{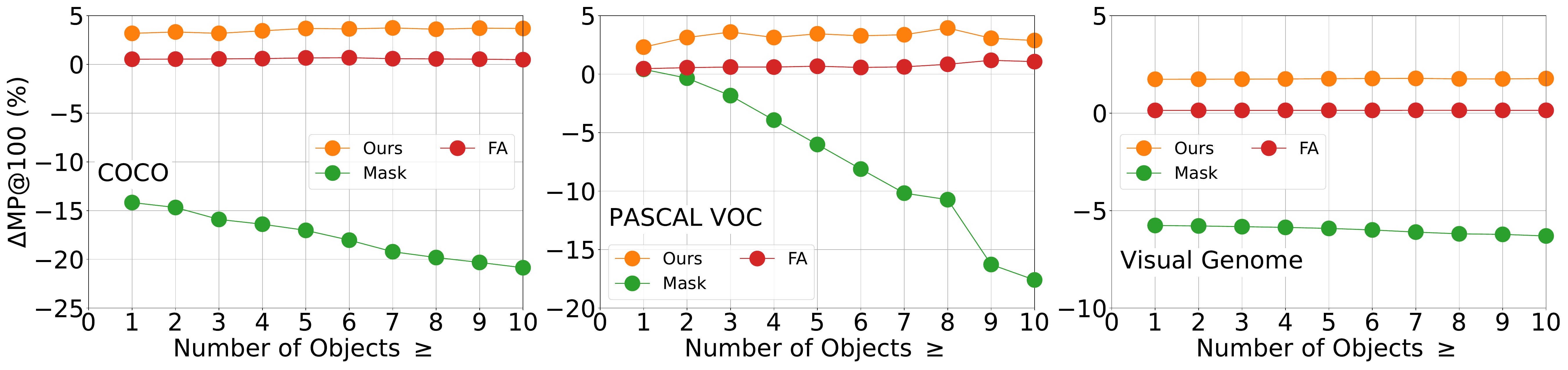}
    \caption{DINOv2 \textit{base}}
    \label{fig:objects-dinov2-base}
    \vspace{0.3cm}
  \end{subfigure}
  \begin{subfigure}{1.0\linewidth}
    \centering
    \includegraphics[width=1.0\linewidth]{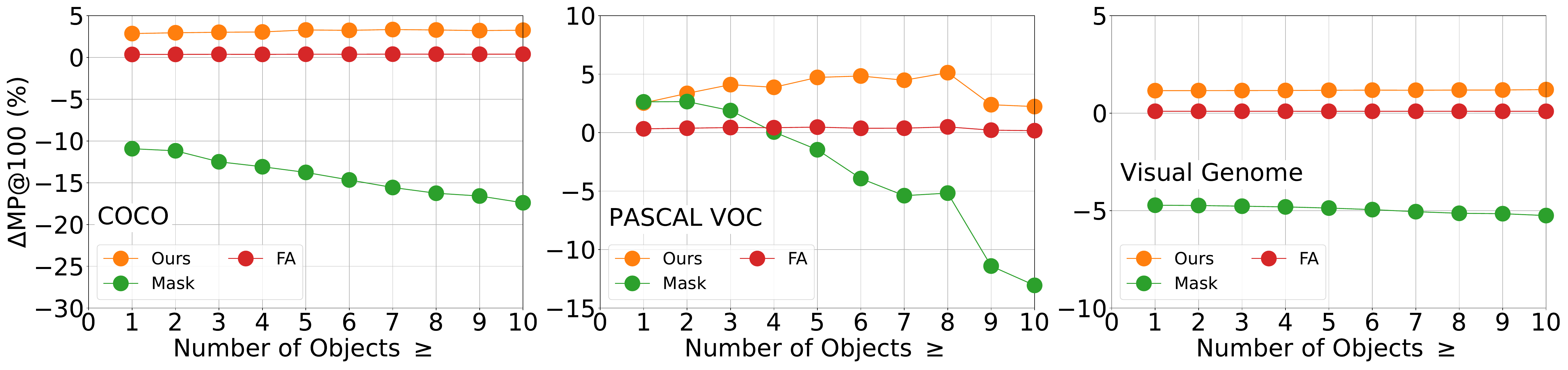}
    \caption{DINOv2 \textit{large}}
    \label{fig:objects-dinov2-large}
    \vspace{0.3cm}
  \end{subfigure}
  \begin{subfigure}{1.0\linewidth}
    \centering
    \includegraphics[width=1.0\linewidth]{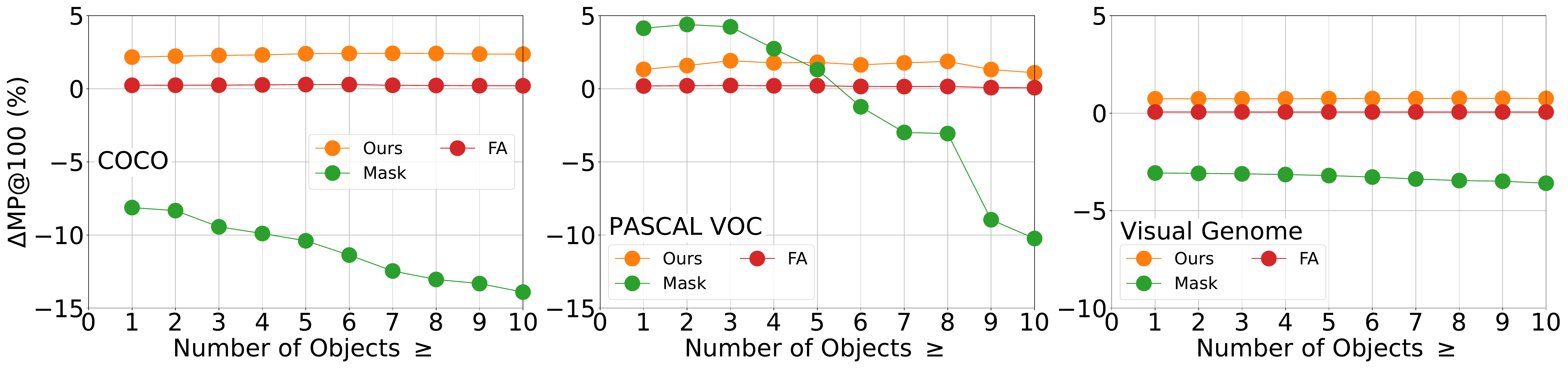}
    \caption{DINOv2 \textit{giant}}
    \label{fig:objects-dinov2-giant}
  \end{subfigure}
  \caption{Relative performance to CBIR. \Crefrange{fig:objects-clip-base}{fig:objects-dinov2-giant} show the results for different models, respectively. The horizontal axis represents that only query images with the number of contained objects equal to or greater than that value are taken into account.}
  \label{fig:objects-allmodel}
\end{figure}

Here, we present an extended analysis on the relationship between the performance of methods and the number of objects in query images. \cref{fig:objects-allmodel} illustrates the relative performance of the methods with respect to CBIR, where we consider only query images with the number of contained objects equal to or greater than the values on the horizontal axis. This result shows that, except for the DINOv2 \textit{small} model, our method demonstrates substantial enhancements in $\mathrm{MP@}100$ even though the number of objects increases. On the other hand, the \textit{Mask} method consistently exhibits lower performance compared to CBIR as the number of objects increases. In the case of DINOv2 \textit{large} or \textit{giant} for the PASCAL VOC, the \textit{Mask} method outperforms our method when considering all the queries including single-object ones. However, our method outperforms the \textit{Mask} method in both cases of DINOv2 \textit{large} with two or more objects and DINOv2 \textit{giant} with five or more objects, which demonstrates the effectiveness of our method in image retrieval containing many objects.

\subsection{Visual Prompt Noise Analysis: Extended Results}
\label{subsec:promptnoise}
In this study, we examine the influence of noise in visual prompts on the effectiveness of our proposed method when comparing to existing methods. Note that users typically do not generate perfect prompts, necessitating the ability of a prompt-driven method to tolerate some level of noise. To simulate this, we introduce noise into the \textit{Box} prompts by randomly shifting and resizing as described in \cref{sec:detailspromptnoise}. The findings, as depicted in \cref{tab:noisyresult}, demonstrate that our method's accuracies remain consistently stable even in the presence of prompt noise. This suggests that our method effectively handles imperfect prompts due to its perception matching mechanism. \textit{FA} also performs relatively robust in our experiments due to its attention blending operation, although the accuracies in general are lower than our method. However, \textit{Mask's} accuracies deteriorate when applying prompt noise, indicating the inherent limitation in image alteration methods.

\begin{table}[h]
  \begin{center}
  \resizebox{1.0\columnwidth}{!}{
  \begin{tabular}{|c|c|ccccccc|}
  \hline
  Dataset & Model & \multicolumn{7}{|c|}{Method and Noise}\\
          & Size  & CBIR\cite{darcet2023vision} & Mask & Mask-N & FA\cite{falip} & FA-N & Ours & Ours-N\\
  \hline
  \multirow{4}{*}{COCO}
  & small & 54.8 & 35.1 & 31.4 & 55.3 & 55.2 & 54.9 & 54.8\\
  & base  & 57.4 & 43.2 & 38.5 & 57.9 & 57.8 & 60.6 & 59.6\\
  & large & 58.4 & 47.5 & 42.1 & 58.8 & 58.7 & 61.3 & 60.6\\
  & giant & 58.5 & 50.4 & 44.6 & 58.8 & 58.7 & 60.7 & 60.4\\
  \hline
  \multirow{4}{*}{\shortstack{PASCAL\\ VOC}}
  & small & 77.2 & 72.7 & 65.4 & 77.8 & 77.6 & 77.4 & 77.1\\
  & base  & 78.6 & 79.0 & 70.5 & 79.0 & 78.9 & 80.9 & 79.8\\
  & large & 77.8 & 80.4 & 72.1 & 78.1 & 78.0 & 80.3 & 79.7\\
  & giant & 78.3 & 82.4 & 74.1 & 78.5 & 78.4 & 79.6 & 79.4\\
  \hline
  \multirow{4}{*}{\shortstack{Visual\\ Genome}}
  & small & 30.1 & 20.5 & 19.0 & 30.2 & 30.2 & 30.1 & 30.0\\
  & base  & 29.6 & 23.9 & 21.8 & 29.8 & 29.7 & 31.4 & 30.9\\
  & large & 29.1 & 24.3 & 21.9 & 29.1 & 29.1 & 30.2 & 29.9\\
  & giant & 29.1 & 26.1 & 22.9 & 29.2 & 29.2 & 29.9 & 29.7\\
  \hline
  \end{tabular}
  }
  \end{center}
  \caption{FOIR results with noisy prompts. Method names end with \textit{-N} represent noisy prompts (Model: DINOv2, Metric: $\mathrm{MP@}100~(\mathrm{\%})$).}
  \label{tab:noisyresult}
\end{table}

\subsection{ROI Attention Strategy Analysis}
\label{subsec:attentionstrategy}
In our proposed method, we utilize the \textit{Sum} operation of attention values within the region of interest (ROI) defined by the user-defined prompt to compute the ROI attention for each head. These ROI attentions are then used to determine the selected heads. Alternatively, the \textit{Max} operation can be employed to compute the ROI attention by identifying the patch with the highest value in the ROI. We conducted an ablation study to compare the performance of these two strategies for ROI attention computation. The results presented in \cref{tab:attentionstrategyresult} consistently demonstrate that our method, which employs the \textit{Sum} strategy, achieves superior performance across multiple datasets.

\begin{table}[h]
  \begin{center}
  \begin{tabular}{|c|ccc|}
  \hline
  Dataset &      & \multicolumn{2}{c|}{ROI Attention Strategy}\\
          & CBIR\cite{darcet2023vision} & Max & Sum (ours)\\
  \hline
  COCO          & 58.4 & 60.3 & {\bf 61.3}\\
  \hline
  PASCAL VOC    & 77.8 & 79.2 & {\bf 80.3}\\
  \hline
  Visual Genome & 29.1 & 29.7 & {\bf 30.2}\\

  \hline
  \end{tabular}
  \end{center}
  \caption{FOIR results with different ROI attention computation strategies (Model: DINOv2 \textit{large}, Metric: $\mathrm{MP@}100~(\mathrm{\%})$).}
  \label{tab:attentionstrategyresult}
\end{table}

\subsection{Head Selection Strategy Analysis}
\label{subsec:headstrategy}
In this study, we investigate various strategies for head selection mechanisms. Our method is inspired by Ref.~\cite{nicolicioiu2023learning}, where head selection is performed prior to the output linear projection layer of the MHA module, and the output of the selected heads is multiplied by a scaling factor. It is important to note that alternative head selection strategies exist. For instance, Ref.~\cite{gong2021pay} applies head selection after the output linear projection layer without the use of a scaling factor. Ref.~\cite{pmlr-v162-wu22c} performs head selection before the output linear projection layer, also without using a scaling factor. Additionally, Refs.~\cite{Meng_2022_CVPR,MHSTbib} replaces the attention matrix of selected heads with an identity matrix, which we refer to as the identity type. 

In our evaluation, we consider strategies related to the position of head selection and the inclusion of the scaling factor. In \cref{tab:headstrategyresult}, we denote \textit{Before} and \textit{After} to indicate the position of the head selection operation relative to the output linear projection layer. The inclusion of the scaling factor is denoted as \textit{Scale}, and the identity type is denoted as \textit{Identity}. From the results presented in \cref{tab:headstrategyresult}, our approach (\textit{Before+Scale}) demonstrates the highest accuracy among various strategies. This ablation analysis highlights the significance of both the position of head selection and the scaling factor in enhancing the performance of image retrieval.

\begin{table}[h]
  \begin{center}
  \resizebox{1.0\columnwidth}{!}{
  \begin{tabular}{|c|ccccc|}
  \hline
  Dataset & \multicolumn{5}{|c|}{Head Selection Strategy}\\
          & Identity & After & Before & After+Scale & Before+Scale (ours)\\
  \hline
  COCO          & 59.5 & 58.3 & 59.8 & 57.9 & {\bf 61.3}\\
  \hline
  PASCAL VOC    & 75.5 & 77.2 & 78.7 & 76.9 & {\bf 80.3}\\
  \hline
  Visual Genome & 29.0 & 28.8 & 29.3 & 28.8 & {\bf 30.2}\\
  \hline
  \end{tabular}
  }
  \end{center}
  \caption{Comparisons of head selection strategies. \textit{Before} and \textit{After} indicate the position of head selection operation in relative to output linear projection layer. \textit{Scale} represents the inclusion of scaling factor. \textit{Identity} denotes the identity matrix replacement method (Model: DINOv2 \textit{large}, Metric: $\mathrm{MP@}100~(\mathrm{\%})$).}
  \label{tab:headstrategyresult}
\end{table}

\subsection{Attention Manipulation Strategy Analysis}
\label{subsec:manipulationstrategy}
In this section, we present an additional study on the attention manipulation strategy in the FOIR task. We create a comparative method called \textit{Attention Mask}, where instead of selecting attention heads, we employ the visual prompt to mask the attentions in the final layer of the ViT model. The results, presented in \cref{tab:manipulationstrategyresult}, demonstrate that the \textit{Attention Mask} approach generally outperforms the previous work of \textit{FA} method. However, our proposed method, PHS, still achieves superior performance compared to \textit{Attention Mask}. Nonetheless, it is worth highlighting that the application of the attention mask in the attention mechanism ensures a more stable performance, avoiding the potential instability that may arise when directly applying the mask to the input image, as shown in the result of \textit{Mask}.

\begin{table}[h]
  \begin{center}
  \resizebox{1.0\columnwidth}{!}{
  \begin{tabular}{|c|ccccc|}
  \hline
  Dataset &      &      & \multicolumn{3}{c|}{ROI Attention Strategy}\\
          & CBIR\cite{darcet2023vision} & Mask & FA\cite{falip} & Attention Mask & ours\\
  \hline
  COCO          & 58.4 & 47.5       & 58.8 & 59.9 & {\bf 61.3}\\
  \hline
  PASCAL VOC    & 77.8 & {\bf 80.4} & 78.1 & 78.6 & 80.3\\
  \hline
  Visual Genome & 29.1 & 24.3       & 29.1 & 29.5 & {\bf 30.2}\\

  \hline
  \end{tabular}
  }
  \end{center}
  \caption{FOIR results with different attention manipulation strategies (Model: DINOv2 \textit{large}, Metric: $\mathrm{MP@}100~(\mathrm{\%})$).}
  \label{tab:manipulationstrategyresult}
\end{table}

\subsection{PHS as a Noise Reduction Technique}
\label{subsec:irq}
We conduct an additional study to investigate the potential of our method as a noise reduction technique. In this study, we set up image retrieval by image-region-as-query (IRQ) query format using a crop-based preprocessing technique. Here, we disregard the preprocessing error associated with cropping by utilizing the bounding box labels provided in the datasets as our box prompt. We crop the query images based on the box prompt and resize them to meet the input requirements of the ViT model. We assume that the resulting cropped and resized images contain the necessary information for the retrieval task. In this particular scenario, our method is employed not to select the essential attention, but rather to exclude any undesired noisy attention. To achieve this, we set the value of $h_{\mathrm{on}}$ to $h-1$, effectively deactivating a single head corresponding to the undesired noise. The results, as depicted in \cref{tab:irqresult}, indicate that our method achieves superior performance compared to the baseline approach for larger DINOv2 models, although by a slight margin. However, for smaller models, our method performs slightly worse, consistent with our observations in the FOIR results. Nevertheless, it is noteworthy that our method consistently outperforms the baseline approach for all cases involving CLIP models. These outcomes suggest the promising potential of our method as an effective technique for attenuating attention noise in images.

\begin{table}[h]
  \begin{center}
  \resizebox{1.0\columnwidth}{!}{
  \begin{tabular}{|c|c|cc|cc|}
  \hline
  Dataset & Model & \multicolumn{2}{|c|}{DINOv2} & \multicolumn{2}{|c|}{CLIP}\\
          & Size  & CBIR\cite{darcet2023vision} & Ours & CBIR\cite{clip} & Ours\\
  \hline
  \multirow{4}{*}{COCO}
            &small & {\bf 60.0} & 58.5       & -    & -   \\
            &base  & {\bf 66.7} & 66.5       & 45.2 & {\bf 46.0}\\
            &large & 67.3       & {\bf 67.4} & 52.3 & {\bf 52.5}\\
            &giant & 68.7       & {\bf 68.8} & -    & -   \\
  \hline
  \multirow{4}{*}{PASCAL VOC} 
            &small & {\bf 86.3} & 85.3       & -    & -   \\  
            &base  & 86.8       & {\bf 86.9} & 76.4 & {\bf 77.0}\\
            &large & 83.9       & {\bf 84.1} & 77.8 & {\bf 77.9}\\
            &giant & 84.0       & {\bf 84.1} & -    & -   \\
  \hline
  \multirow{4}{*}{Visual Genome} 
            &small & {\bf 34.2} & 33.4       & -    & -   \\  
            &base  & {\bf 34.7} & 34.6       & 24.0 & {\bf 24.4}\\
            &large & {\bf 33.4} & {\bf 33.4} & 25.7 & {\bf 25.8}\\
            &giant & 34.5       & {\bf 34.6} & -    & -   \\
  \hline
  \end{tabular}
  }
  \end{center}
  \caption{PHS as a noise reduction technique (Metric: $\mathrm{MP@}100~(\mathrm{\%})$).}
  \label{tab:irqresult}
\end{table}

\subsection{Visual Analysis with Attention Map: Extended Results}
\label{subsec:visualattfull}
\begin{figure}[h]
  \centering
  \includegraphics[width=\linewidth]{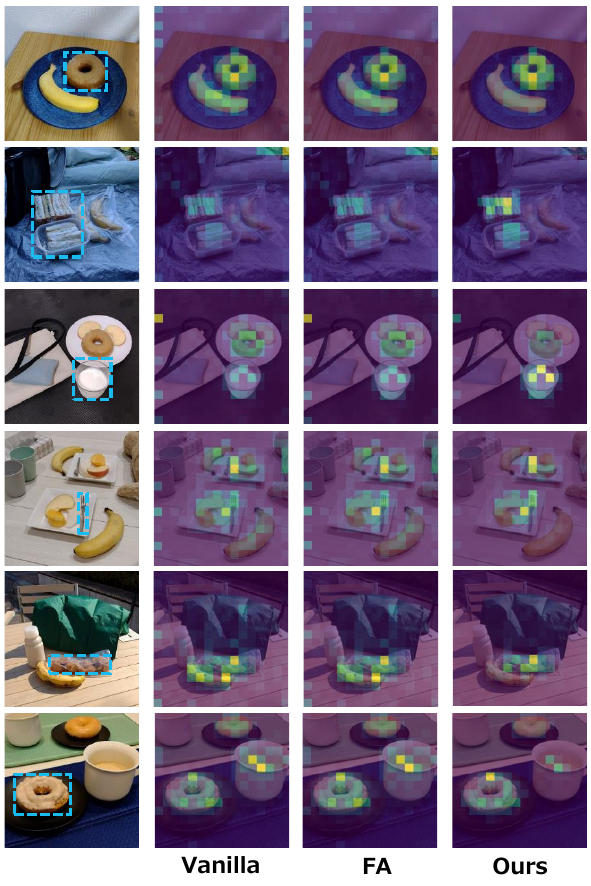}
  \caption{The visualization of attention maps demonstrates that our method performs more intuitively than \textit{Vanilla} and \textit{FA}. Best viewed in color (Model: DINOv2 \textit{giant}).}
  \label{fig:visualizefull}
\end{figure}

Here, we present the extended results of our visualization analysis on attention maps generated in the final layer of the ViT model, after incorporating our proposed method. As depicted in \cref{fig:visualizefull}, our method demonstrates superior intuition in terms of enhanced focus and noise reduction when comparing to \textit{Vanilla} ViT (used in \textit{CBIR}) and \textit{FA}. In contrast, \textit{FA} typically generates attention maps that are comparable to those produced by \textit{Vanilla} ViT, albeit with slightly more concentrated ROI attentions. It is noteworthy that our approach preserves potentially valuable surrounding visual context, which plays a crucial role in reflecting user perception.

\subsection{Visual Analysis with Attention Map Across Multiple Model Sizes}
\label{subsec:visualsizes}
\begin{figure*}[t]
  \centering
  \includegraphics[width=\linewidth]{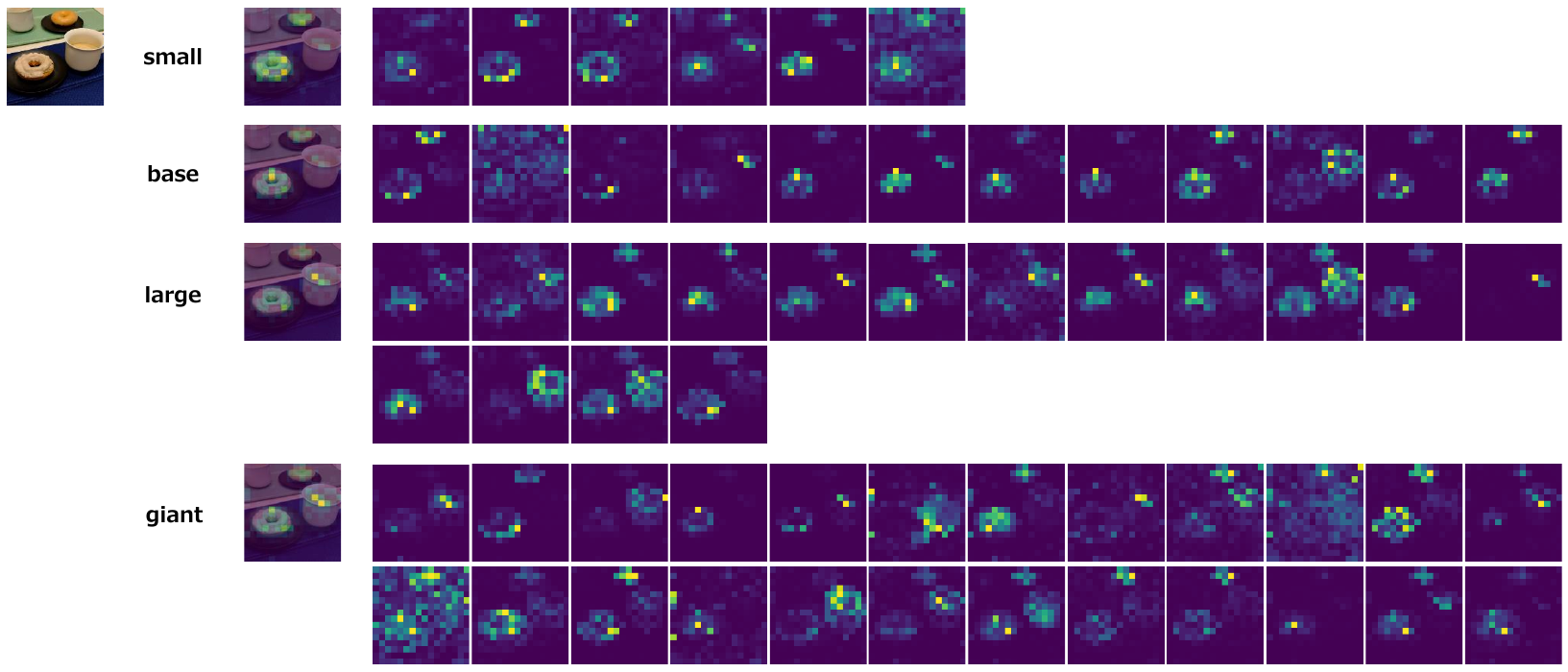}
  \caption{Attention maps visualization across different model sizes for individual attention heads in \textit{Vanilla} ViT. Best viewed in color (Model: DINOv2).}
  \label{fig:attentionmodel}
\end{figure*}

\begin{figure*}[b]
  \centering
  \includegraphics{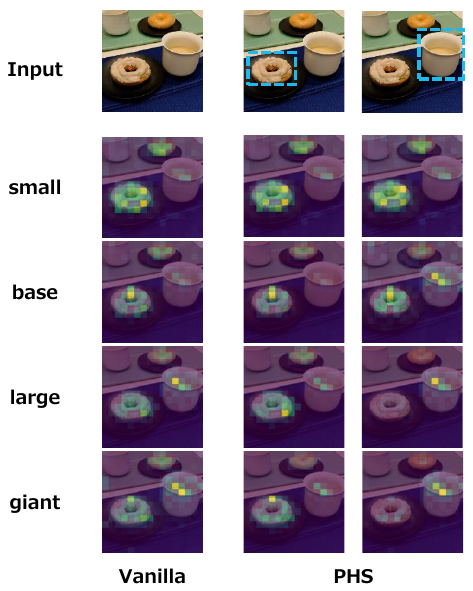}
  \caption{Attention maps visualization across different model sizes when applying our method. Best viewed in color (Model: DINOv2).}
  \label{fig:attentionheadon}
\end{figure*}

In this section, we present a comprehensive visual analysis and investigation of the attention maps generated by \textit{Vanilla} ViT models and our method across different model sizes. \cref{fig:attentionmodel} illustrates attention maps of individual heads in the last layer of \textit{Vanilla} ViT for various model sizes. The \textit{base}, \textit{large}, and \textit{giant} models exhibit distinct differentiations across attention heads, indicating the potential of selecting objects based on attention heads. However, the \textit{small} model displays limited differentiation due to its smaller number of heads. This observation aligns with the overall weaker results obtained with our method on the \textit{small} model in our experiments. When applying our approach, \cref{fig:attentionheadon} demonstrates the remarkable alignment between the attention map, visual prompt, and input image with the \textit{giant} and \textit{large} models. Conversely, the \textit{small} model exhibits noisy attention maps even after applying our proposed method. The \textit{base} model's visual quality is somewhere in between. This observation underscores the limitations of our method when dealing with models that have a smaller number of attention heads.

\section{Metrics used in Performance Evaluations}
\label{sec:metrics}
In this section, we describe the details of the performance metrics used in our experiments. To evaluate the performance of our method, we use Mean Precision at $k$ ($\mathrm{MP@}k$) and Mean Average Precision at $k$ ($\mathrm{MAP@}k$), following Ref.~\cite{eccv2020metriclearning}. 
Let $\mathcal{C}$ be the set of categories of objects.
We assume that each query image $\mathbf{x}_{\mathrm{Q}}$ includes objects $o_1,o_2,\dots,o_{n\left(\mathbf{x}_{\mathrm{Q}}\right)}$.
We define the category of $o_{i}$ as $c\left(o_{i}\right) \in \mathcal{C}$, and the number of objects in category $c$ as $n_{c}\left(\mathbf{x}_{\mathrm{Q}}\right)$.
In our experiments, it is important to note that the correctness of retrieved images depends on the visual prompt, even if the query image is the same.
For a query image $\mathbf{x}_{\mathrm{Q}}$ with a visual prompt for $o_{i}$, 
we consider the $k^\prime$th retrieved image $\mathbf{x}_{k^\prime}$ as \textit{correct} if it contains an object in category $c\left(o_{i}\right)$ and \textit{incorrect} if it does not.
We define the score $S$ for $\mathbf{x}_{k^\prime}$ as follows:
\begin{align}
  S\left(\mathbf{x}_{k^\prime}, \mathbf{x}_{\mathrm{Q}}, o_{i}\right) = 
  \begin{cases}
    1 & \text{if $\mathbf{x}_{k^\prime}$ is correct,}\\
    0 & \text{if $\mathbf{x}_{k^\prime}$ is incorrect.}
  \end{cases}
\end{align}
Then, $\mathrm{MP@}k$ are calculated by
\begin{align}
  &\mathrm{\tilde{P}@}k\left(\mathbf{x}_{\mathrm{Q}},o_{i}\right)
  =\frac{1}{k}\sum_{1\leq k^{\prime}\leq k} S\left(\mathbf{x}_{k^\prime}, \mathbf{x}_{\mathrm{Q}}, o_{i}\right),\\
  \label{eq:patk}
  &\mathrm{P@}k\left(c\right) = \frac{1}{|\mathcal{I}_{\mathrm{Q},c}|}
  \sum_{\mathbf{x}_{\mathrm{Q}}\in\mathcal{I}_{\mathrm{Q},c}}
  \sum_{i:c\left(o_{i}\right)=c}
  \frac{\mathrm{\tilde{P}@}k\left(\mathbf{x}_{\mathrm{Q}},o_{i}\right)}{n_{c}\left(\mathbf{x}_{\mathrm{Q}}\right)},\\
  \label{eq:mpatk}
  &\mathrm{MP@}k = 
  \frac{1}{|\mathcal{C}|}\sum_{c\in\mathcal{C}}\mathrm{P@}k\left(c\right),
\end{align}
where $\mathcal{I}_{\mathrm{Q},c}$ is the set of query images that include objects in category $c$.
$\mathrm{\tilde{P}@}k$ is the proportion of correct images in the top-$k$ ones for each visual prompt based query. 
$\mathrm{P@}k$ is the average of $\mathrm{\tilde{P}@}k$ over visual prompt based queries for a fixed $c$, 
and $\mathrm{MP@}k$ is the average of $\mathrm{P@}k$ over $\mathcal{C}$.
$\mathrm{MAP@}k$ are calculated by
\begin{align}
  &\widetilde{\mathrm{AP}}\mathrm{@}k\left(\mathbf{x}_{\mathrm{Q}},o_{i}\right) = 
  \frac{1}{|\mathcal{K}|}\sum_{k^{\prime}\in \mathcal{K}}
  \mathrm{\tilde{P}@}k^{\prime}\left(\mathbf{x}_{\mathrm{Q}},o_{i}\right),\\
  &\mathcal{K} = \left\{k^{\prime}\in\left\{1,2,\dots,k\right\}\mid
  \text{$\mathbf{x}_{k^\prime}$ is correct}\right\},\\
  \label{eq:apatk}
  &\mathrm{AP@}k\left(c\right)
  = \frac{1}{|\mathcal{I}_{\mathrm{Q},c}|}
  \sum_{\mathbf{x}_{\mathrm{Q}}\in\mathcal{I}_{\mathrm{Q},c}} 
  \sum_{i:c\left(o_{i}\right)=c}
  \frac{\widetilde{\mathrm{AP}}\mathrm{@}k\left(\mathbf{x}_{\mathrm{Q}},c\right)}{n_{c}\left(\mathbf{x}_{\mathrm{Q}}\right)},\\    
  \label{eq:mapatk}
  &\mathrm{MAP@}k = 
  \frac{1}{|\mathcal{C}|}\sum_{c\in\mathcal{C}}\mathrm{AP@}k\left(c\right).
\end{align}

$\mathrm{AP@}k$ and $\mathrm{MAP@}k$ are metrics that value the higher-ranking images more than $\mathrm{P@}k$ and $\mathrm{MP@}k$. In this paper, we employ $\mathrm{MP@}k$ and $\mathrm{MAP@}k$ as our performance metrics and set $k$ to 100.

\section{Details of Visual Prompt Noise}
\label{sec:detailspromptnoise}
In our experiments, visual prompt noise is added in the following way.
For an object in each original query image, the noiseless box prompt is specified by the positions of the upper left corner $(x_{0}, y_{0})$ and the lower right corner $(x_{1}, y_{1})$ of the box.
For the visual prompt with noise, we change $(x_{0}, y_{0})$ and $(x_{1}, y_{1})$ to $(\tilde{x}_{0}, \tilde{y}_{0})$ and $(\tilde{x}_{1}, \tilde{y}_{1})$ randomly as follows:
\begin{align}
  (\tilde{x}_{0}, \tilde{y}_{0}) &= (x_{0}, y_{0}) + (\tilde{c}_{x}, \tilde{c}_{y}) - (\tilde{l}_{x}, \tilde{l}_{y}),\\
  (\tilde{x}_{1}, \tilde{y}_{1}) &= (x_{1}, y_{1}) + (\tilde{c}_{x}, \tilde{c}_{y}) + (\tilde{l}_{x}, \tilde{l}_{y}),
\end{align}
where $\tilde{c}_{x}$, $\tilde{c}_{y}$, $\tilde{l}_{x}$, and $\tilde{l}_{y}$ are sampled from the discrete uniform distribution over $[-m, m]$ respectively.
The box prompt is shifted by $(\tilde{c}_{x}, \tilde{c}_{y})$ and resized by $(\tilde{l}_{x}, \tilde{l}_{y})$.
In all our experiments with noise, we set $m$ to $40$ pixels, which is roughly 7.6\% of image width and height in average for COCO, 9.4\% for PASCAL VOC, and 9.0\% for Visual Genome.

\section{Licence info}
\label{sec:license}

\Cref{tab:license} shows the license info of image used in Fig.~7 of the paper.

\begin{table}[h]
   \centering
   \resizebox{1.0\linewidth}{!}{
   \begin{tabular}{|c|c|}
      \hline
      Image id & 563470 \\
      \hline
      URL & \url{ http://farm4.staticflickr.com/3370/3518451715_596120fc59_z.jpg } \\
      \hline
      \multirow{2}{*}{License}
      & CC BY-NC-SA 2.0 DEED \\
      & \url{http://creativecommons.org/licenses/by-nc-sa/2.0/} \\
      \hline
   \end{tabular}
   }
   \caption{License info of image in Fig.~7 of the paper.}
   \label{tab:license}
\end{table}

\end{document}